\pdfoutput=1

\documentclass[11pt]{article}

\usepackage{algorithm}
\usepackage{algpseudocode}
\usepackage{amsmath}
\usepackage{amssymb}
\usepackage{url}
\algdef{SE}[TRY]{Try}{EndTry}{\textbf{try}}{\textbf{end try}}
\algtext*{EndTry}

\algrenewcommand\algorithmicrequire{\textbf{Require:}}
\algrenewcommand\algorithmicensure{\textbf{Ensure:}}

\usepackage[final]{acl}

\usepackage{times}
\usepackage{latexsym}

\usepackage[T1]{fontenc}

\usepackage[utf8]{inputenc}

\usepackage{microtype}

\usepackage{inconsolata}

\usepackage{graphicx}
\graphicspath{{../figures/}}

\usepackage{booktabs}
\usepackage{multicol}
\usepackage{multirow}
\usepackage{tcolorbox}
\usepackage{colortbl}
\usepackage{util}
\usepackage{xspace}

\definecolor{NavyBlue}{RGB}{0,51,102}

\lstdefinestyle{python}{
    language=Python,
    basicstyle=\fontsize{7}{9.5}\ttfamily,
    keywordstyle=\color{blue},
    commentstyle=\color{gray},
    stringstyle=\color{black},
    showstringspaces=false,
    breaklines=true,
    breakindent=0pt,
    breakatwhitespace=false,
    escapeinside={(*@}{@*)}
}

\lstdefinestyle{plain}{
    basicstyle=\fontsize{7}{9.5}\ttfamily,
    keywordstyle=\color{blue},
    commentstyle=\color{gray},
    stringstyle=\color{green},
    showstringspaces=false,
    breaklines=true,
    breakatwhitespace=false,
    breakindent=0pt,
    escapeinside={(*@}{@*)}
}

\newcommand{\dataset}{\textsc{PGPS9K}\xspace}
\newcommand{\pal}{\textsc{PAL}\xspace}

\newcommand{\CoT}{CoT\xspace}

\newcommand{\improve}[1]{{\scriptsize \color{red} \textbf{(+#1)}}}
\definecolor{darkgreen}{RGB}{0, 100, 0} %
\newcommand{\dec}[1]{\textcolor{darkgreen}{\scriptsize{(-#1)}}} %

\title{Reactivating Test-Time Scaling for Plane Geometry Problem Solving}

\author{
    Xiaoqiang Kang$^{1,2}$, 
    Shengen Wu$^{4,5}$, 
    Maizhen Ning$^{1,2}$, 
    Xiaobo Jin$^1$, \\
    \textbf{Kaizhu Huang}$^3$, 
    \textbf{Yutao Yue}$^4$, 
    \textbf{Xiaowei Huang}$^2$, 
    \textbf{Qiufeng Wang}$^{1,}$\thanks{Corresponding author.} \\
    $^1$School of Advanced Technology, Xi'an Jiaotong-Liverpool University\\
    $^2$Department of Computer Science, University of Liverpool \\ 
    $^3$Digital Innovation Research Center, Duke Kunshan University \\
    $^4$The Hong Kong University of Science and Technology (Guangzhou), $^5$Hithink Research \\
    \texttt{Xiaoqiang.Kang23@student.xjtlu.edu.cn, Qiufeng.Wang@xjtlu.edu.cn}
}

\begin{document}
\maketitle
\begin{abstract}
Plane geometry problem (PGP) solving has become a critical benchmark for multimodal reasoning because it requires accurate visual perception and precise multi-step symbolic deduction. 
Although test-time scaling (TTS) has demonstrated remarkable success in general mathematical reasoning, it fails to scale effectively under the symbolic-program paradigm for plane geometry.
We identify two key obstacles: limited reasoning diversity induced by rigid symbolic programs and insufficient explicit visual grounding before symbolic deduction.
To address these issues, we propose \textbf{M}ulti-\textbf{T}race \textbf{S}ynthesis (MTS), which converts each symbolic program into heterogeneous reasoning traces, including executable Python scripts and \CoT-augmented variants. 
We further propose \textbf{P}erception-\textbf{A}ugmented (PA) training, which parses diagrams into structured semantic clauses before deduction, and \textbf{C}onsensus-\textbf{G}uided \textbf{M}ulti-\textbf{T}race \textbf{E}nsemble (CG-MTE) for efficient self-adaptive inference. 
Experiments on three geometry benchmarks show that our method consistently improves PGP-solving across model scales and achieves strong performance against both general-purpose MLLMs and specialized geometry solvers.
Under test-time scaling, CG-MTE achieves a comparable accuracy to high-budget self-consistency while reducing sampling cost by up to 8$\times$.
Code and data are publicly available at \url{https://github.com/Jason8Kang/ReTTS-PGPS}.
\end{abstract}

\section{Introduction}
Plane geometry problem (PGP) solving is a fundamental benchmark for multimodal reasoning \cite{cho-etal-2026-plane}.
Previous research has advanced PGP through formal symbolic systems, neuro-symbolic solvers, and, more recently, Multimodal Large Language Models (MLLMs) \cite{zhaoGeometryProblemSolving2025}.
However, PGP-solving remains challenging for current MLLMs, because it requires both accurate recognition of visual entities and relations and precise multi-step deduction \cite{2025geoxxia}. 
Concurrently, test-time scaling (TTS) techniques \cite{snell2025scaling}, particularly self-consistency (SC) \cite{wang2023selfconsistency}, have emerged as a standard paradigm for improving the reasoning performance of MLLMs by aggregating multiple sampled pathways.
It remains unclear, however, whether this scaling behavior transfers effectively to PGP.

\begin{figure}[t]
    \centering
    \includegraphics[width=\linewidth]{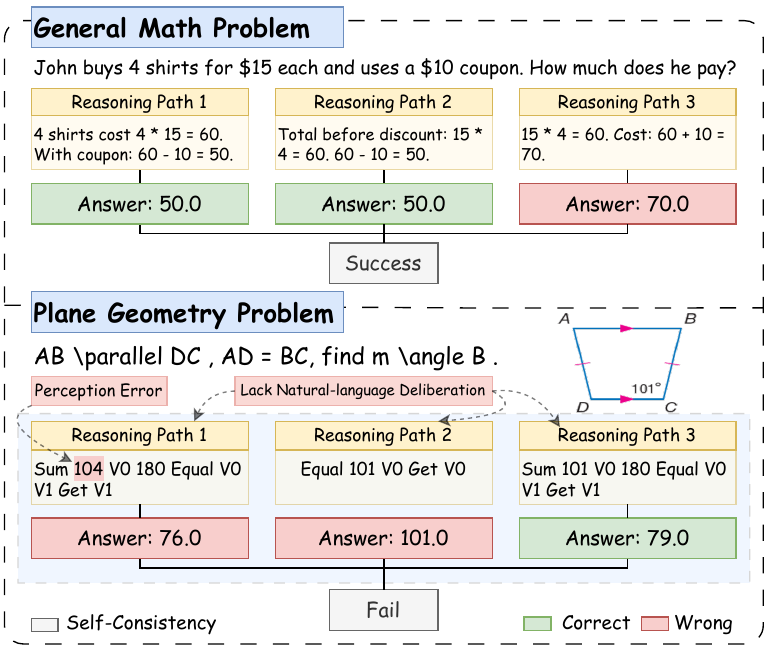}
    \caption{Comparison of self-consistency in general math and plane geometry problems.}
    \vspace{-4mm}
    \label{fig:tts_comparison}
\end{figure}

Our empirical observations reveal that TTS provides only limited benefits under the symbolic program prediction paradigm for PGP. 
As Figure~\ref{fig:tts_comparison} illustrates, while SC in general mathematics benefits from exploring various reasoning paths that often converge to the correct answer, sampled trajectories on PGP often fail to reach a majority consensus.
We attribute this failure to two main factors. 
First, \textbf{reasoning diversity is limited}. Existing geometry datasets rely heavily on rigid symbolic programs. These concise formal languages lack the natural-language deliberation required to explore a broad reasoning space. %
Second, \textbf{a perceptual gap remains}. As shown in Figure~\ref{fig:tts_comparison} (reasoning path 1),  the model misreads visual signals, for example, by interpreting $101^{\circ}$ as $104^{\circ}$. These perception errors subsequently lead to incorrect deductions.
Consequently, the correct program is often unique, and repeated sampling tends to produce either the same correct program or multiple inconsistent incorrect ones, which weakens the effectiveness of SC.

To address the constrained reasoning space, we propose a \textbf{M}ulti-\textbf{T}race \textbf{S}ynthesis (MTS) framework that expands each symbolic program into semantically aligned but heterogeneous reasoning traces. %
Specifically, rigid programs are transformed into Python scripts. Both programs and scripts are further augmented with Chain-of-Thought (\CoT) rationales in natural language, reintroducing the intermediate deliberation. %
Fine-tuning on the mixture of original programs and synthesized traces enables models to generate diverse reasoning pathways, thereby reactivating test-time scaling.

To bridge the gap between visual input and formal deduction, we introduce \textbf{P}erception-\textbf{A}ugmented (PA) training. This training paradigm optimizes the capability of the model to first parse diagrams into structured semantic clauses before performing reasoning. 
It strengthens the visual-to-symbolic alignment and provides a more reliable foundation for subsequent deduction.

While MTS and PA training establish the basis for effective TTS, such scaling also exposes a practical challenge: computational efficiency. 
Naively increasing the sampling budget can introduce redundancy, especially for easy problems. 
To address this, we propose a self-adaptive inference strategy, \textbf{C}onsensus-\textbf{G}uided \textbf{M}ulti-\textbf{T}race Ensemble (CG-MTE). 
Whereas standard MTE aggregates all trajectories from every trace type regardless of difficulty, CG-MTE checks cross-trace consensus at shallow decoding depths and expands the sampling budget only when disagreement persists.
This strategy allocates additional sampling on uncertain instances, yielding a better accuracy-cost trade-off.

Based on the MTS framework, we construct MTS-All datasets for three geometry datasets, namely PGPS9K-All, Geometry3K-All, and GeoQA-All.
In our experiments, we fine-tune the Qwen-VL family and evaluate on all three benchmarks.
Experimental results demonstrate that PA training on the corresponding MTS-All datasets consistently surpasses baselines, including strong general-purpose MLLMs and specialized geometry solvers.
On PGPS9K, our fine-tuned Qwen3-VL-8B achieves 71.2\% accuracy under greedy decoding.
With test-time scaling, self-consistency with beam search further improves accuracy to 76.7\% using 40 samples.
In comparison, CG-MTE reaches a comparable accuracy of 76.0\% with an average sampling number of only 4.89, substantially improving the accuracy--compute trade-off.

Our main contributions are summarized as follows:
\textbf{(1)} We propose a program-seeded MTS framework that transforms symbolic programs into heterogeneous reasoning traces, which enables effective test-time scaling for PGP-solving.
\textbf{(2)} We introduce PA training that grounds symbolic deduction in structured semantic clauses parsed from diagrams.
\textbf{(3)} We propose CG-MTE, a self-adaptive inference strategy, improving the Pareto trade-off between accuracy and computational cost.
\textbf{(4)} Based on MTS, we construct MTS-All datasets for PGPS9K, Geometry3K, and GeoQA and achieve strong performance on all benchmarks.

\section{Related Work}
\paragraph{PGP Solving} demands the integration of multimodal comprehension and multi-step logical deduction. 
Early neural-symbolic methods, such as Inter-GPS \citep{lu2021intergps} and GeoDRL~\citep{peng-etal-2023-geodrl}, parse diagrams and text into formal languages and use symbolic solvers to perform deduction. 
Though interpretable and verifiable, they are sensitive to parsing errors as symbolic deduction relies on precise problem formalization.
Another line of work focuses on neural solvers, including NGS \citep{chen2021geoqa}, Geoformer \citep{chen2022unigeo}, PGPSNet \citep{Zhang2023PGPS}, and LANS \citep{li-etal-2024-lans}, which directly generate solution programs from multimodal problems.

Recent MLLM-based methods further improve geometry reasoning through visual-language pretraining and logical reasoning fine-tuning~\citep{gaoGLLaVA2023,pan2025enhancing,cheng2025geouni}. 
Another emerging direction is program-aided reasoning, where models generate executable programs to support symbolic computation, as exemplified by the Program-Aided Language (\pal) paradigm~\citep{gao2023pal}. 
AlphaGeometry \citep{trinhSolvingOlympiadGeometry2024} exemplifies this synthesis by combining a neural model for intuitive auxiliary constructions with a formal deduction engine to reach Olympiad-level proficiency. Concurrently, other studies explore program-based geometry reasoning~\citep{ning2025gns, 2025geoxxia}. 
Different from these solver-centric approaches, our work focuses on improving PGP-solving through diverse reasoning traces and adaptive TTS.

\paragraph{Data synthesis for PGP} addresses the shortage of high-quality annotations, which is a main bottleneck in training MLLMs.
Early dataset construction relied on manual effort, establishing foundational but inherently limited resources \citep{lu2021intergps,chen2021geoqa,haoPGDP5KDiagramParsing2022,Zhang2023PGPS,zhangFormalGeoExtensibleFormalized2024a}. 
UniGeo \citep{chen2022unigeo} and G-LLaVA \citep{gaoGLLaVA2023} scaled data by empirical augmentation, but the diversity of reasoning traces is limited.
To ensure mathematical rigor beyond simple data scaling, works such as AlphaGeometry~\citep{trinhSolvingOlympiadGeometry2024} and GeoFM~\cite{zhangGeoFMEnhancingGeometric2025} synthesize data through symbolic deduction and formal constraints. 
However, their synthesized problems often diverge linguistically from human-authored problems, limiting their effectiveness for natural language alignment.
Consequently, recent research has focused on improving the quality of reasoning traces rather than simply data generation~\cite{zhaoAchievingOlympiaLevelGeometry2025,wangConciseGeometricDescription2026,chenMilestonesOutcomeUnlocking2026}. 
GeoThought \citep{shiGeoThoughtDatasetEnhancing2025} employs a teacher MLLM to generate new problems with \CoT and improves process-level supervision through rejection sampling and consensus verification.
TR-CoT injects theorem knowledge into the generation process, yet it remains susceptible to hallucinations~\cite{trcot}.
In contrast, our MTS framework does not synthesize new problems. 
Instead, it derives multiple semantically aligned traces from verified formal programs, improving reasoning diversity while preserving the rigor of the underlying symbolic reasoning.

\begin{figure*}[!t]
    \centering
    \includegraphics[width=\linewidth]{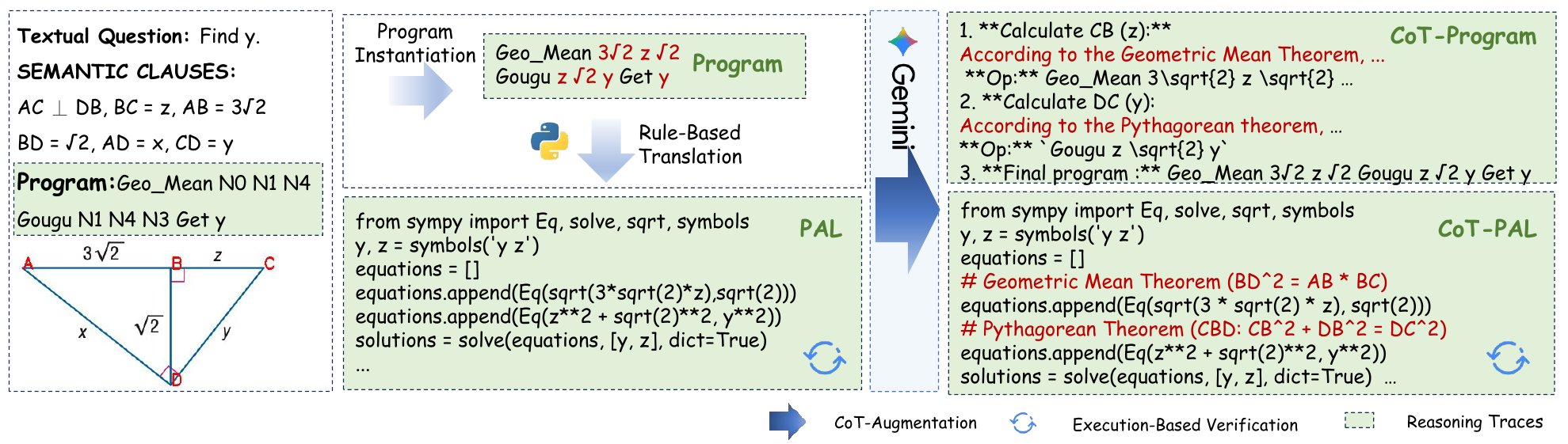}
    \caption{Overview of the multi-trace synthesis pipeline. Starting from a solution program, we first instantiate variables and translate it into a verified executable \pal script via rule-based translation and execution-based verification. We then augment the instantiated program and the verified \pal script with natural-language rationales to obtain \CoT-Program and \CoT-\pal, respectively.}
    \vspace{-4mm}
    \label{fig:example}
\end{figure*}

\section{Methodology}
\label{sec:method}

\subsection{Problem Formulation}
\label{sec:problem_formulation}
Formally, we define a geometry problem instance as a tuple $\mathcal{I} = (D, Q, S, P, A)$, which consists of a diagram image $D$, a textual question $Q$, semantic clauses $S$, a solution program $P$ (i.e., a sequence of symbolic reasoning steps), and a numeric answer $A$. 
Semantic clauses describe explicit geometric relations or measurements, such as $AC \perp DB$, $AB = 3\sqrt{2}, BC = z$, as illustrated in Figure~\ref{fig:example}. 
Our objective is to train an MLLM that takes $D$ and $Q$ as input, first predicts semantic clauses $S$, and then generates a reasoning trajectory $T$ that leads to the correct answer $A$. 

\paragraph{Formal Solution Program.}
\label{sec:formal_language}
Our method relies on a domain-specific language for rigorous geometric deduction. 
Formally, a solution program is a sequence of program steps 
$P = \langle s_1, \ldots, s_T \rangle$, 
where each step $s_t = (\mathsf{op}_t, \mathsf{args}_t)$ applies a predefined geometric operator $\mathsf{op}_t \in \mathcal{O}$ to a list of operands, including problem variables (\texttt{N}), process variable (\texttt{V}), and constants (\texttt{C}). 
As illustrated in Figure~\ref{fig:example}, a program step $s_t = \texttt{Gougu}(\texttt{N1}, \texttt{N4}, \texttt{N3})$ denotes the application of the Pythagorean theorem to known side lengths \texttt{N1} and \texttt{N4} to calculate the hypotenuse \texttt{N3}.
Detailed definitions of the operators and operands are in Appendix~\ref{sec:formal_details}.

\noindent\textbf{Program Instantiation.}
Executing a symbolic program requires an auxiliary mapping that binds problem variables to the corresponding numerical values.
To streamline execution, we render the program self-contained by (i) substituting \texttt{N} with values, and (ii) converting \texttt{C} into constants.
For example, as shown in Figure~\ref{fig:example}, the program \texttt{Geo\_Mean N0 N1 N4 Gougu N1 N4 N3 Get y} becomes \texttt{Geo\_Mean $3\sqrt{2}$ z $\sqrt{2}$ Gougu z $\sqrt{2}$ y Get y}  after instantiation.
This eliminates the need to construct an auxiliary mapping during preprocessing and provide such a mapping at execution time.

\subsection{Multi-Trace Synthesis (MTS)}
\label{sec:reasoning_traces}

To diversify PGP-solving strategies, we expand each instantiated program into three reasoning traces across two dimensions: (i) \textbf{format transformation}, converting symbolic programs to executable Python scripts; and (ii) \textbf{rationale augmentation}, integrating \CoT into both formats. 

\subsubsection{Program-to-\pal Conversion}
\label{sec:pal}
Following \citet{gao2023pal}, we adopt the Program-Aided Language (\pal) paradigm, where a language model solves problems by generating executable scripts (typically Python). In our setting, \pal refers to Python scripts that solve geometric problems.
We generate these scripts through a rigorous pipeline consisting of a rule-based translation followed by execution-based verification.

\noindent\textbf{Rule-Based Translation.}
This stage converts the instantiated program into a standardized Python template:
(1) Environment Initialization: set up necessary imports and symbolic variables.
(2) Equation Formulation: map each program step to an algebraic equation. 
(3) Equation Solving: append a solver block to resolve the system of equations.
(4) Post-processing: filter invalid roots (e.g., negative lengths) to yield the numerical answer.

\noindent\textbf{Execution-Based Verification.}
To ensure script correctness, we execute it in a sandbox with a timeout constraint.
The script is valid only if it runs successfully without errors and returns a result $\hat{a}$ that matches the ground-truth answer $a$ within a relative tolerance ($\epsilon = 0.001$). 

\subsubsection{\CoT Augmentation}
\label{sec:cot}
While the instantiated programs and derived \pal scripts are executable, they do not explicitly articulate the intermediate geometric reasoning.
To bridge this gap, we introduce an MLLM-based \CoT augmentation stage that rewrites them into rationale-enriched variants. This process yields the following \CoT-style traces.

\noindent\textbf{\CoT-\pal} enhances the standard \pal scripts by adding explicit natural-language rationales before each equation. 
As illustrated in Figure~\ref{fig:example}, the geometric rationale (e.g., \textit{``Geometric Mean Theorem ($BD^2 = AB * BC$)''}) is first articulated before formulating the corresponding equation.
To improve the reliability of the generated traces, we implement a generate-and-verify pipeline. 
Every transformed script undergoes the same execution-based verification described in Section~\ref{sec:pal}. 
If execution fails, the error message is returned to the MLLM as feedback, and the full loop is repeated up to three times until success.
The prompts for \CoT-\pal conversion and bug fixing are detailed in Tables~\ref{tab:prompt_cotpal} and \ref{tab:prompt_cotpal_fix} in Appendix \ref{apd:prompt_for_reasoning}.

\noindent\textbf{\CoT-Program} rewrites the solution program into a structured, step-by-step natural-language explanation aligned with each program step, following prior work~\cite{Yang_Li_Li_Lin_Sun_Ding}.
As shown in Figure~\ref{fig:example}, each step explains the application of geometric principles before the program step. 
Table~\ref{tab:prompt_cotprogram} in Appendix \ref{apd:prompt_for_reasoning} contains the complete prompt used for \CoT-Program transformation.
Because \CoT-Program augments the program with natural-language explanations, the resulting trace is not directly executable as a whole, and thus execution-based verification is not applicable.
During inference, we parse the program from the generated \CoT-Program and execute it to obtain the answer.

\begin{figure*}[thbp]
    \centering
    \includegraphics[width=\linewidth]{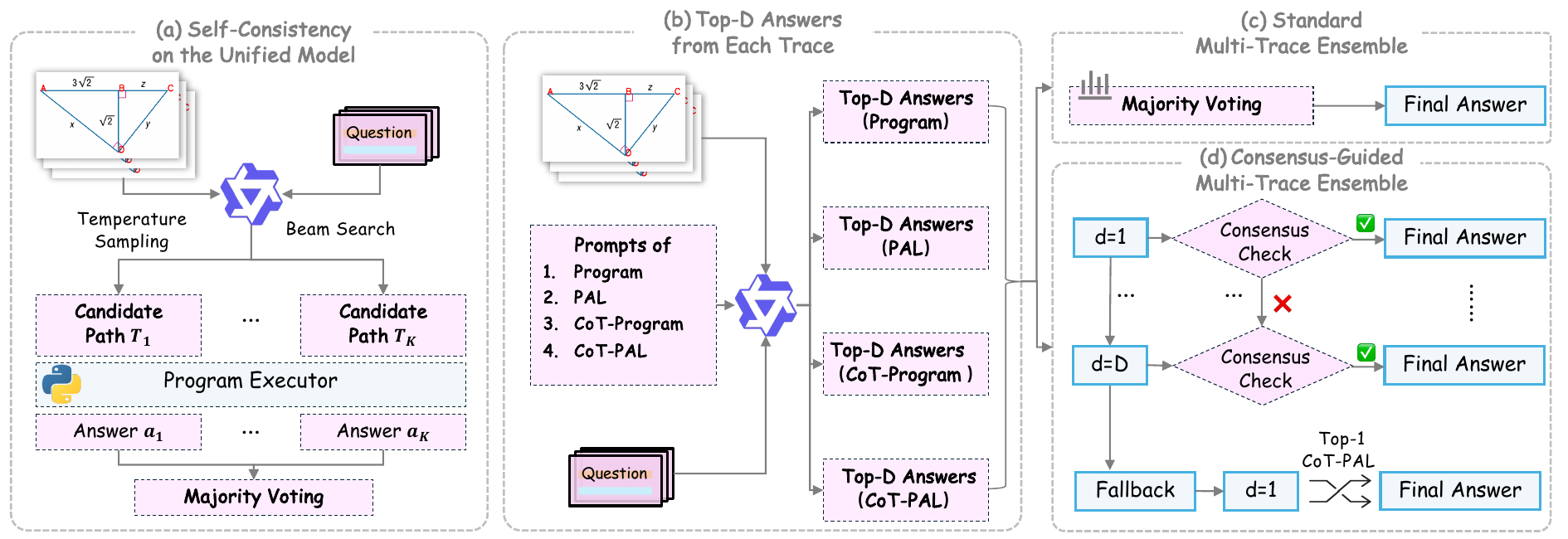}
    \caption{Comparison of test-time scaling strategies. (a) Self-consistency on the unified model. (b) Multi-trace inference with trace-specific instructions. (c) Standard MTE. (d) CG-MTE expands depth until consensus.}
    \vspace{-4mm}
    \label{fig:inference_framework}
\end{figure*}

\subsection{Perception-Augmented Training}
\label{sec:training_stage}
Solving a diagram-based geometry problem requires both diagram understanding and symbolic reasoning.
Recent multimodal reasoning frameworks such as LLaVA-CoT~\citep{Xu_2025_LLava_CoT} and Insight-V~\citep{dong2025insight} emphasize explicit visual grounding prior to complex reasoning. 
Inspired by this paradigm, we incorporate a perception step to bridge the visual-to-symbolic gap in geometry reasoning.
Specifically, we formulate the problem-solving process as a sequential, perception-augmented pipeline, termed \textbf{Perception-Augmented (PA) Training}:
(i) \textbf{Perception}, where the MLLM parses the diagram $D$ into explicit semantic clauses $S$; and
(ii) \textbf{Reasoning}, which conditions on the perceived clauses $S$ to generate an executable reasoning trace $T$.

We optimize this integrated process by maximizing the joint log-likelihood $P_\theta(S, T \mid D, Q)$:
\begin{equation}
\scalebox{0.92}{$
\begin{aligned}
\mathcal{L}
&= - \log P_\theta(S, T \mid D, Q) \\
&= - \log \big[ P_\theta(S \mid D, Q)\cdot P_\theta(T \mid S, D, Q) \big] \\
&= \underbrace{-\log P_\theta(S \mid D, Q)}_{\mathcal{L}_{\text{perc}}}  \underbrace{-\log P_\theta(T \mid S, D, Q)}_{\mathcal{L}_{\text{reason}}} 
\end{aligned}
$}
\label{eq:joint_loss_derivation}
\end{equation}

\noindent where $S=(S_1,\dots,S_{|S|})$ denotes the parsed semantic clauses, and 
$T=(T_1,\dots,T_{|T|})$ denotes the reasoning trace.
The perception loss $\mathcal{L}_{\text{perc}}$ encourages accurate parsing of diagram semantics, while the reasoning loss $\mathcal{L}_{\text{reason}}$ trains the model to generate the reasoning trace conditioned on structured visual evidence.
Both terms are optimized autoregressively with standard cross-entropy.

\subsection{Inference: Test-Time Scaling}
\label{sec:inference}
Building on diverse PGP-solving strategies, we explore two strategies: (1) self-consistency (SC) on a unified model, and (2) a multi-trace ensemble that aggregates various reasoning trace types.

\subsubsection{Self-Consistency on the Unified Model}
For the model trained on a mixture of reasoning traces, the unified model can implicitly select among different reasoning strategies at test time without prompt control. 
We employ standard SC \citep{wang2023selfconsistency} to aggregate multiple candidate reasoning paths, as illustrated in Figure~\ref{fig:inference_framework}(a).
Given an input $x=(D, Q)$, we generate $K$ candidate paths $\{T_1, \dots, T_K\}$ via beam search or temperature sampling. Each path is executed to produce an answer $a_i = \textsc{Exec}(T_i)$.

The final answer $a^*$ is determined by majority voting over the executed results of all $K$ paths:
\begin{equation}
\scalebox{0.8}{$
a^* = \mathrm{mode}(\{a_i\}_{i=1}^K) = \arg\max_{v \in \mathcal{V}} \sum_{i=1}^K \mathbb{I}(a_i = v),
$}
\label{eq:majority_voting}
\end{equation}
where $\mathcal{V} = \{a_1, \dots, a_K\}$ represents the set of candidate answers and $\mathbb{I}(\cdot)$ is the indicator function.

\subsubsection{Multi-Trace Ensemble (MTE)}
\label{sec:consensus_guided_voting}
To leverage the diversity of reasoning traces, we generate $V$ distinct traces by prompting the same fine-tuned model with trace-specific instructions. Detailed prompts are provided in Appendix~\ref{apd:mte_prompts}. 
For each trace type $j\in\{1, \dots, V\}$, we apply beam search to obtain the top-$D$ paths.
These trajectories are executed to produce an answer list
\begin{equation}
\mathcal{A}_j(x)=[a_{j,1}(x),\dots,a_{j,D}(x)].
\end{equation}
We consider two aggregation mechanisms displayed in Figure~\ref{fig:inference_framework}(c) and (d).

\paragraph{Standard MTE.}
A straightforward baseline pools all $VD$ executed answers and performs majority voting as in Equation \ref{eq:majority_voting}.
While effective, this approach is computationally expensive, as it requires generating the maximum number of samples for each problem, regardless of difficulty.

\paragraph{Consensus-Guided MTE (CG-MTE).}
To improve computational efficiency, we perform consensus checking progressively from shallow to deep and terminate as soon as a unique mode emerges.
Concretely, we iterate over the depth $d\in\{1,\dots,D\}$ and apply the following four steps:
\textbf{(1) Step-wise Ensemble:}
At depth $d$, pool the top-$d$ answers from all trace types into an answer list 
$\mathbb{S}_d(x)=\bigcup_{j=1}^{V}\{a_{j,1}(x),\dots,a_{j,d}(x)\}$.
\textbf{(2) Consensus Check:}
Compute $v_d^*(x)=\mathrm{mode}(\mathbb{S}_d(x))$.
If $v_d^*(x)$ exists, we terminate and output $a^* = v_d^*(x)$.
\textbf{(3) Depth Expansion:}
If $v_d^*(x)$ does not exist, increase the depth $d\leftarrow d+1$ and repeat Steps (1)--(2), until $d=D$.
\textbf{(4) Fallback:}
If $v_D^*(x)$ still does not exist, we return the top-1 \CoT-\pal answer as the fallback.
Therefore, CG-MTE does not introduce an additional consensus threshold hyperparameter. 
It serves only as an empirical stopping rule rather than a correctness certificate, because the trace outputs may exhibit correlated errors.

\paragraph{Compute efficiency.}
We quantify inference cost by the average sampling number (ASN), the number of generated candidates per instance.
For the Standard MTE, the sample cost is fixed:
\begin{equation}
ASN_{\text{Std}}(D)=VD\ .
\end{equation}
For the CG-MTE, let $\mathcal{T}$ denote the test set and let $d_x$ denote the termination depth for instance $x \in \mathcal{T}$.
The sampling cost is
\begin{equation}
ASN_{\text{CG}}(D)=\frac{1}{|\mathcal{T}|}\sum_{x\in\mathcal{T}}Vd_x=\frac{V}{|\mathcal{T}|}\sum_{x\in\mathcal{T}}d_x,
\end{equation}
which enables early termination when consensus is reached at shallow depths, reducing sampling cost.

\section{Experiments}
\label{sec:exp}

\paragraph{Dataset and Metrics.}
We evaluate our method on three widely used geometry problem-solving benchmarks: \dataset~\citep{Zhang2023PGPS}, Geometry3K~\citep{lu2021intergps}, and GeoQA~\citep{chen2021geoqa}. 
Since GeoQA does not provide semantic-clause annotations required by PA training, we manually annotate them following the same format as \dataset. 
After applying the MTS framework, each symbolic program is expanded into four reasoning-trace variants. We denote the MTS-All datasets as PGPS9K-All, Geometry3K-All, and GeoQA-All, which contain approximately 32.1K, 33.7K, and 13.9K training instances, respectively.
Detailed MTS construction procedures and dataset statistics are provided in Appendix~\ref{apd:MTS_Framework}.
Following prior work, we report answer accuracy (top-1 match). An answer is considered correct if the predicted numerical value matches the ground truth within a relative tolerance of $\epsilon = 10^{-3}$.

\paragraph{Compared Systems.}
We compare our method against three categories of systems:
(i) \textbf{General-purpose MLLMs}: frontier and open-source multimodal large language models including GPT-4V~\cite{openai2023gpt4}, Claude 3.5 Sonnet~\cite{anthropic2024claude3}, and Qwen2.5-VL~\cite{bai2025qwen25vl}.
(ii) \textbf{Neural Geometry Solvers}: end-to-end geometry reasoning systems trained specifically for diagram understanding and program prediction, including NGS~\cite{chen2021geoqa}, Geoformer~\cite{chen2022unigeo}, PGPSNet~\cite{Zhang2023PGPS}, PGPSNet-v2-S~\cite{zhang2024fuse}, LANS~\cite{li-etal-2024-lans}, and GeoX~\cite{2025geoxxia}.
(iii) \textbf{Neural-symbolic Geometry Solvers}: systems that integrate neural perception with symbolic geometric reasoning, including InterGPS~\cite{lu2021intergps}, GeoDRL~\cite{peng-etal-2023-geodrl}, and Pi-GPS~\cite{wang2025pigps}.
We fine-tune three backbones: Qwen2-VL-2B-Instruct, Qwen2.5-VL-3B-Instruct, and Qwen3-VL-8B-Instruct.
For brevity, we denote them as 2B, 3B, and 8B, respectively. 
To verify that improvements are not specific to Qwen-VL, additional experiments on InternVL3.5-2B and InternVL3.5-8B are provided in Appendix~\ref{apd:internvl_results}.

\paragraph{Implementation Details.}
For multi-trace synthesis, we use Gemini-2.5-Flash \citep{gemini25flash} to generate \CoT rationales.
Unless otherwise specified, we report main results with greedy decoding. 
For each benchmark, our model is trained with PA training on its corresponding MTS-All dataset.
For the test-time scaling analysis, all models are trained with PA training.
For self-consistency, we use beam search or temperature sampling and set the sampling budget to $K=40$.
Since MTS-All includes four reasoning trace types---program, \pal, \CoT-Program, and \CoT-\pal---we set $V=4$ in all multi-trace inference experiments.
For each trace type, we keep the top-$D$ candidates with $D=10$, yielding the same maximum budget of $VD=40$ for MTE.
Additional training hyperparameters are provided in Appendix~\ref{apd:implementation_details}.

\begin{table}[t]
    \centering
    \small
    \resizebox{\linewidth}{!}{
    \begin{tabular}{lccc}
        \toprule
        \textbf{Model} & \textbf{PGPS9K} & \textbf{Geometry3K} & \textbf{GeoQA} \\
        \midrule

        \rowcolor{gray!10}
        \multicolumn{4}{c}{\textit{General-purpose MLLMs}} \\

        GPT-4V & 33.3 & 34.8 & 43.4 \\
        Claude 3.5 Sonnet & 27.6 & 32.0 & 49.2 \\

        Qwen2.5-VL-7B & 39.4 & 35.8 & 46.2 \\
        Qwen2.5-VL-72B & 53.3 & 50.5 & 55.5 \\

        \midrule

        \rowcolor{gray!10}
        \multicolumn{4}{c}{\textit{Neural Geometry Solvers}} \\

        NGS & 34.1 & 35.3 & 46.3 \\
        Geoformer & 35.6 & 36.8 & 49.1 \\
        GeoX & 52.7 & 58.6 & 54.9 \\
        PGPSNet & 62.7 & 65.0 & -- \\
        PGPSNet-v2-S & 60.3 & 65.2 & -- \\
        LANS & 66.7 & 72.1 & -- \\
        
        \midrule

        \rowcolor{gray!10}
        \multicolumn{4}{c}{\textit{Neural-symbolic Geometry Solvers}} \\

        InterGPS (Diagram GT) & 59.8 & 64.2 & -- \\
        GeoDRL & 55.6 & 57.9 & -- \\
        Pi-GPS & 61.4 & 70.6 & -- \\

        \midrule

        \rowcolor{gray!10}
        \multicolumn{4}{c}{\textit{Fine-tuned Models}} \\

        \multicolumn{4}{l}{\quad \textit{Baseline: Direct Program Prediction on Solution Program}} \\

        Qwen3-VL-8B & 58.6 & 65.4 & 60.9 \\
        Qwen2.5-VL-3B & 46.7 & 53.1 & 53.4 \\
        Qwen2-VL-2B & 43.4 & 41.4 & 47.1 \\

        \multicolumn{4}{l}{\quad \textit{Ours: PA Training on MTS-All}} \\

        Qwen3-VL-8B & \textbf{71.2} \improve{12.6} & \textbf{74.5} \improve{9.1} & \textbf{67.2} \improve{6.3} \\
        Qwen2.5-VL-3B & 58.3 \improve{11.6} & 64.7 \improve{11.6} & 58.7 \improve{5.3} \\
        Qwen2-VL-2B & 50.8 \improve{7.4} & 52.0 \improve{10.6} & 53.5 \improve{6.4} \\

        \bottomrule
    \end{tabular}
    }

    \caption{
    Performance comparison on three benchmarks.
    All metrics are reported as answer accuracy (\%).
    }

    \vspace{-4mm}
    \label{tab:main_results}
\end{table}

\subsection{Main Results}
\label{sec:main_results}

Table~\ref{tab:main_results} reports the comparative results on three benchmarks. Our key findings are as follows:

\paragraph{Significantly Enhanced Reasoning Capabilities.} 
PA training on MTS-All consistently outperforms direct program prediction across all model scales and benchmarks.
On PGPS9K, it yields absolute gains of 7.4\%, 11.6\%, and 12.6\% for the 2B, 3B, and 8B backbones, respectively, with similar gains on Geometry3K and GeoQA.
These results show that structured perception and heterogeneous reasoning traces jointly improve geometric reasoning.

\paragraph{Strong Performance Across Benchmarks.} 
Our fine-tuned Qwen3-VL-8B achieves 71.2\%, 74.5\%, and 67.2\% accuracy on PGPS9K, Geometry3K, and GeoQA, respectively.
Compared with general-purpose MLLMs, it substantially outperforms Qwen2.5-VL-72B by 17.9\%, 24.0\%, and 11.7\% on the three benchmarks.
Compared with specialized geometry solvers, it also achieves the best results on PGPS9K and Geometry3K, surpassing the strongest compared solver, LANS, by 4.5\% and 2.4\%, respectively.

\section{Analysis}
\label{sec:analysis}

\subsection{Ablation Studies}
\label{sec:ablation_study}

\paragraph{Effectiveness of PA Training.}
The efficacy of the PA training paradigm is underscored by the comparative results in Table~\ref{tab:ablation_modules}.
Removing the perception step and reverting to direct program prediction leads to consistent accuracy drops across all model scales, namely 5.3\%, 6.5\%, and 5.8\% on the 2B, 3B, and 8B backbones, respectively.  
This degradation shows that explicit semantic parsing bridges the visual-to-symbolic gap by grounding geometric reasoning in accurate visual signals. 
Qualitative case studies in Appendix~\ref{apd:perception_cases} show how the perception step reduces erroneous symbolic deductions caused by diagram misinterpretation.

\begin{table}[h]
    \centering
    \setlength{\tabcolsep}{6pt}
    \resizebox{\linewidth}{!}{%
    \begin{tabular}{lcccc}
        \toprule
        \textbf{Setting} & \textbf{Data Size} & \textbf{2B} & \textbf{3B} & \textbf{8B} \\
        \midrule
        \textbf{PA Training on PGPS9K-All} & 32K & \textbf{50.8} & \textbf{58.3} & \textbf{71.2} \\
        \quad w/o PA Training & 32K  & 45.5 \dec{5.3} & 51.8 \dec{6.5} & 65.4 \dec{5.8} \\
        \quad w/o PGPS9K-MTS & 32K & 46.7 \dec{4.1} & 54.4 \dec{3.9} & 67.6 \dec{3.6}\\
        \bottomrule
    \end{tabular}
    }
    \caption{Ablation study of PA training and the synthesized PGPS9K-MTS. Numbers in \textcolor{darkgreen}{green} indicate absolute accuracy drops.}
    \vspace{-4mm}
    \label{tab:ablation_modules}
\end{table}

\paragraph{Effectiveness of Diverse Reasoning Traces.} 
To isolate the effect of reasoning diversity from data scale, we introduce a data-size-matched control by repeating the original symbolic programs of PGPS9K to 32K, matching the size of PGPS9K-All.
Removing the synthesized reasoning traces (w/o PGPS9K-MTS) leads to a consistent performance degradation of approximately 4\%, as shown in Table~\ref{tab:ablation_modules}.
To further disentangle the contributions of different reasoning formats, we analyze single-trace variants in Table~\ref{tab:ablation_traces}. 
Compared to single-trace settings, merging all traces into PGPS9K-All achieves the best performance, suggesting that diverse reasoning traces provide complementary strategies and improve generalization.

\begin{table}[h]
    \centering
    \small 
    \setlength{\tabcolsep}{4pt} 
    \resizebox{0.9\linewidth}{!}{
    \begin{tabular}{lccccc} 
        \toprule 
        \textbf{Training Data} & \textbf{2B} & \textbf{3B} & \textbf{8B} & \textbf{Len} & \textbf{Correctness} \\ 
        \midrule 
        \multicolumn{6}{l}{\textit{Single Trace (PGPS9K) }} \\ 
        \quad Program & 46.2 & 54.2 & 67.5 & -- & -- \\ 
        \quad \CoT-Program & 46.0& 54.5& 67.9 & 5.1K & 93.0\% \\ 
        \quad \pal & 46.7 & 54.4& 67.4 & -- & -- \\ 
        \quad \CoT-\pal & \textbf{47.2} & \textbf{55.4}& \textbf{68.3} & 0.9K & 98.5\% \\ 
        \midrule 
        \multicolumn{6}{l}{\textit{Multi-Trace (PGPS9K-All)}} \\ 
        \quad PGPS9K-All & \textbf{50.8} & \textbf{58.3} & \textbf{71.2} \\ 
        \bottomrule 
    \end{tabular} 
} 
\caption{Performance comparison across individual traces and multi-trace mixtures. \textbf{Correctness} and \textbf{Len} denote human-verified correctness and prompt length.} 
\vspace{-4mm}
\label{tab:ablation_traces} 
\end{table}

\begin{figure*}[thbp] %
    \centering
    \includegraphics[width=\linewidth]{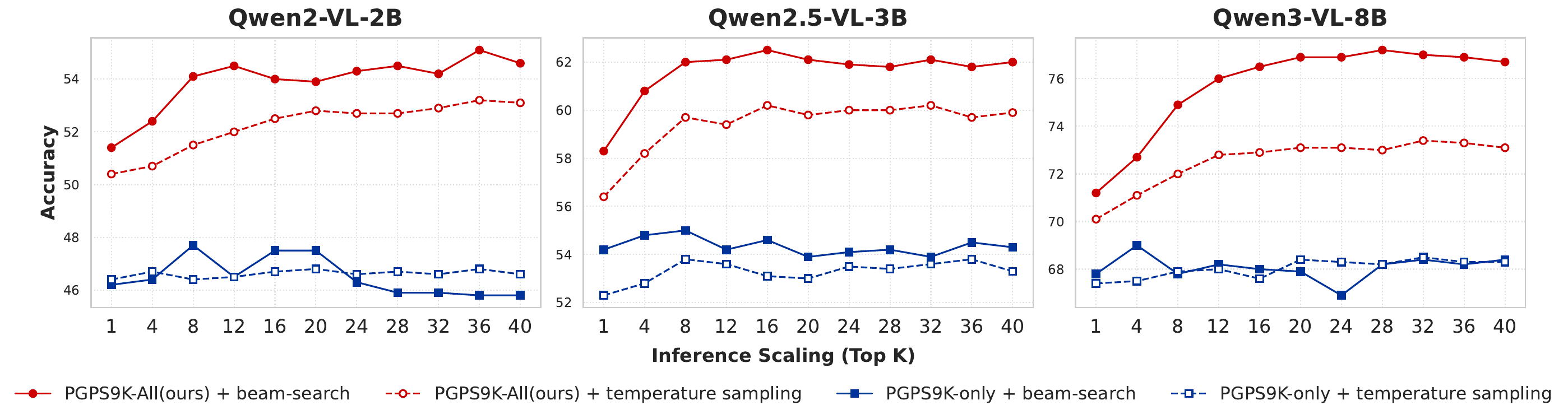}
    \caption{Impact of data diversity on test-time scaling across model sizes. Models trained on PGPS9K-All scale robustly with increasing sample budget, whereas the PGPS9K-only baseline shows limited gains or even degradation.}
    \vspace{-4mm}
    \label{fig:inference_scaling}
\end{figure*}

\paragraph{Superiority of Executable \CoT-\pal.}
\CoT-\pal consistently achieves the best performance across all model scales compared to other single-trace types. As detailed in Table~\ref{tab:ablation_traces}, it achieves accuracies of 47.2\%, 55.4\%, and 68.3\% on the 2B, 3B, and 8B backbones, respectively.
This performance advantage is consistent with the higher reasoning faithfulness of the synthesized traces produced by our translate-then-rewrite pipeline.
To verify this, we conduct a human evaluation on two \CoT-style traces to assess the correctness of the reasoning process. %
The results, summarized in Table~\ref{tab:ablation_traces}, reveal that \CoT-\pal achieves a reasoning correctness rate of 98.5\%, surpassing \CoT-Program by 5.5\%. 
Additional details regarding the human evaluation and the higher training efficiency gains of \CoT-\pal are provided in Appendix~\ref{apd:trace_quality} and Appendix~\ref{apd:training_efficiency}, respectively.

\begin{table}[h]
    \centering
    \small 
    \setlength{\tabcolsep}{4pt} 
    \resizebox{0.90\linewidth}{!}{
    \begin{tabular}{lccc}
    \toprule 
    \textbf{Setting} & \textbf{Tokens (O/M)} & \textbf{2B} & \textbf{3B} \\
    \hline
    MTS-All & 7.2M & \textbf{50.8} & \textbf{58.3} \\
    w/o \CoT-Program & 4.9M / 7.2M & 49.7 \dec{-1.1} & 56.9 \dec{-1.4} \\
    w/o \pal & 5.1M / 7.2M & 48.6 \dec{-2.2} & 56.6 \dec{-1.7} \\
    w/o \CoT-\pal & 5.0M / 7.2M & 47.7 \dec{-3.1} & 55.3 \dec{-3.0} \\
    w/o Program & 6.6M / 7.2M & 48.9 \dec{-1.9} & 56.1 \dec{-2.2} \\
    \bottomrule
    \end{tabular}
}
\caption{Token-controlled leave-one-out ablation of individual reasoning traces on PGPS9K. Tokens (O/M) denote the original and matched training token budgets.}
\vspace{-4mm}
\label{tab:leave-one-out}
\end{table}

\paragraph{Effectiveness of Individual Trace Types.}
While Table~\ref{tab:ablation_traces} compares single-trace and multi-trace settings, it does not reveal whether each individual trace contributes positively to the full mixture. Therefore, we conduct token-controlled leave-one-out ablations on PGPS9K. Specifically, after removing one trace type from MTS-All, we resample the remaining traces to match the MTS-All token budget. As shown in Table~\ref{tab:leave-one-out}, removing any individual trace type consistently decreases performance across both backbones. 
This demonstrates that different trace formats provide complementary reasoning trajectories. In particular, removing \CoT-\pal leads to the largest degradation, suggesting that \CoT-\pal provides a particularly effective reasoning format among the four trace formats. These results also indicate that \CoT-augmented traces do not fully subsume their non-CoT counterparts.

\paragraph{Efficacy of Data Construction Pipeline.} 
\label{sec:ablation_construction} 
We further analyze why \CoT-\pal outperforms \CoT-Program by examining their data construction processes.
\CoT-Program is produced in a single pass, and the resulting traces cannot be validated through code execution.
In contrast, \CoT-\pal is derived through a rigorous three-stage pipeline: (1) rule-based translation preserves the program’s logical skeleton; (2) MLLM rewriting enriches it with natural language rationale; (3) execution-based verification ensures executable correctness.
This decomposition reduces the difficulty of generating rationales and improves faithfulness.
Moreover, constructing \CoT-Program requires providing semantic definitions for all 34 geometric operators as inputs to the MLLM, whereas \CoT-\pal leverages the derived \pal script as input, thereby substantially reducing the prompt length from $5.1$K to $0.9$K tokens.

\subsection{Scaling Laws of Test-Time Compute}
\label{sec:test_time_compute}

\paragraph{Scaling Performance of Self-Consistency.} 
Training on the diverse reasoning traces of PGPS9K-All yields better test-time scaling behavior than the PGPS9K-only baseline, which is trained solely on symbolic programs. As illustrated in Figure~\ref{fig:inference_scaling}, applying self-consistency to the PGPS9K-All model yields consistent performance gains across different inference budgets, whereas the program-only baseline shows negligible improvement or even degradation as the inference budget increases. For the Qwen3-VL-8B model, beam search improves accuracy from $71.2\%$ at top-1 to $76.7\%$ at top-40, while temperature sampling increases it from $69.1\%$ to $73.1\%$. 
These results suggest that diverse reasoning traces can unlock the benefits of TTS in PGP-solving. Similar TTS trends on Geometry3K and GeoQA are provided in Appendix~\ref{apd:TTS_on_3_benchmarks}. 
Furthermore, for the PGPS9K-All model, we observe that beam search consistently outperforms temperature sampling with a more detailed analysis in Appendix~\ref{apd:analysis_of_self_consistency}. 

\begin{figure}[!htbp]
    \centering
    \includegraphics[width=\linewidth]{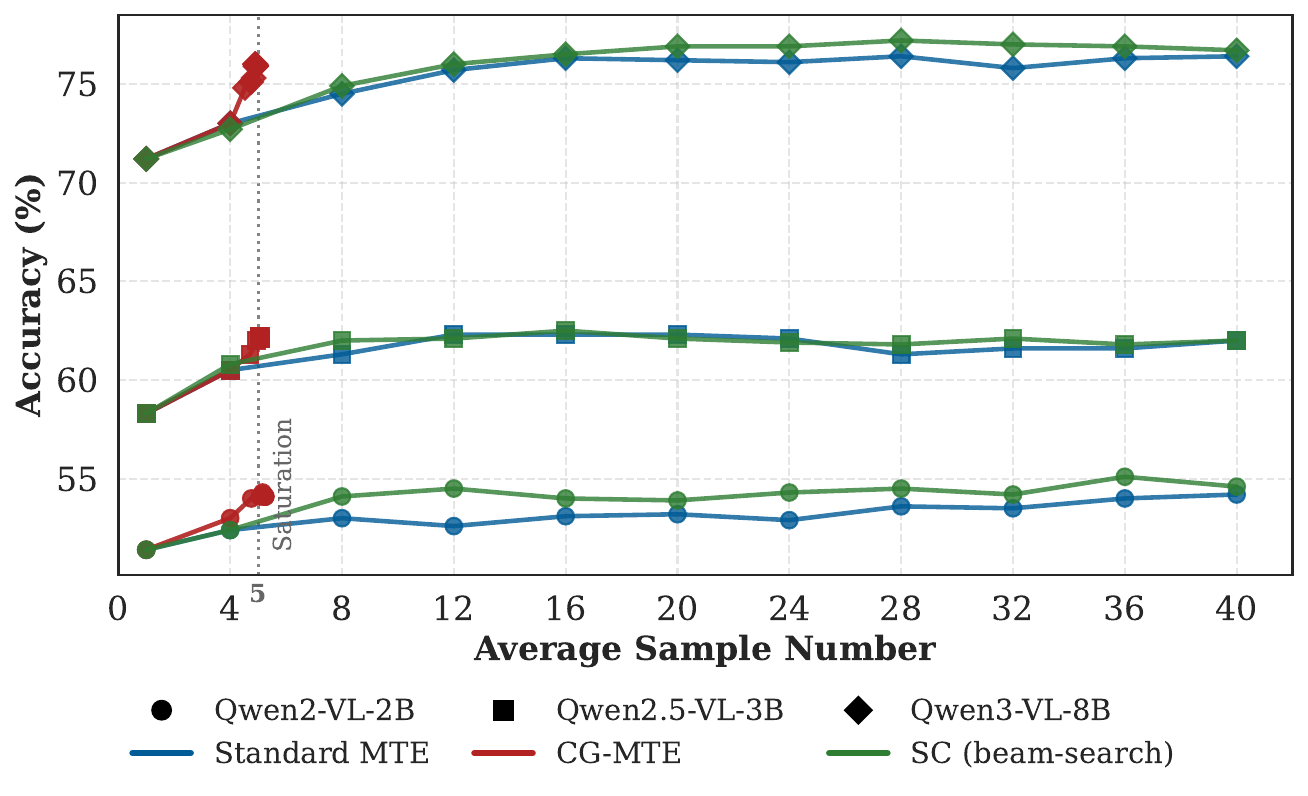}
    \caption{Accuracy--compute trade-off on PGPS9K among Standard MTE, SC  with beam search, and CG-MTE. The dashed line marks the accuracy saturation.}
    \vspace{-4mm}
    \label{fig:inference_scaling_ensemble}
\end{figure}

\paragraph{Pareto Efficiency of CG-MTE.}
We further evaluate the accuracy--compute trade-off of different test-time scaling strategies on PGPS9K.
As shown in Figure~\ref{fig:inference_scaling_ensemble}, we compare SC with beam search, Standard MTE, and CG-MTE under increasing inference budgets.
Both standard MTE and SC exhibit a relatively slow, near-linear scaling trajectory.
By contrast, CG-MTE reaches its performance plateau much earlier, demonstrating better Pareto efficiency across all model scales.
For the Qwen3-VL-8B model, CG-MTE attains $76.0\%$ accuracy with $ASN_{\text{CG}} \approx 5$, as indicated by the vertical dashed line in Figure~\ref{fig:inference_scaling_ensemble}.
This matches the performance of SC with beam search at $N=12$, while using approximately $2.4\times$ fewer samples.
Compared with a high-budget reference setting ($N=40$), CG-MTE remains comparable in accuracy ($76.7\%$ for SC and $76.4\%$ for Standard MTE), while reducing sampling cost by up to $8\times$.
Overall, CG-MTE consistently improves the accuracy–compute trade-off.

\paragraph{Mechanism of Efficient CG-MTE.}
The efficiency of CG-MTE stems from the fact that cross-trace agreement is reached very early for most instances. 
Moreover, early agreement is associated with higher accuracy, whereas cases requiring deeper expansion are typically more uncertain and harder. 
This explains why CG-MTE achieves a better accuracy--compute trade-off than standard MTE. 
Appendix~\ref{apd:consensus_guided_mechanism} provides an in-depth analysis of termination behavior and a representative case.
However, because all four traces are derived from the same symbolic program, their errors may be correlated.
We therefore interpret cross-trace consensus as an empirical early-stopping signal rather than a correctness guarantee.
Appendix~\ref{sec:error-coupling} reports the error-coupling analysis across trace formats.

\paragraph{Comparison with Adaptive Beam-Search SC.} To determine whether the efficiency of CG-MTE comes from grouping and early stopping, we compare it with two adaptive beam-search SC baselines. Both first generate the unified model's top-40 beam candidates, partition them into four groups, and apply the same unique-mode stopping rule and maximum budget as CG-MTE. The rank-split baseline partitions the top-40 candidates from unified beam search into four rank blocks (1–10, 11–20, 21–30, and 31–40), whereas the pattern-split baseline groups candidates according to their generated trace patterns. As shown in Table~\ref{tab:adaptive-beam}, rank split achieves nearly the same ASN as CG-MTE but is 0.8 and 1.0 points less accurate on the 8B and 3B models, respectively. Pattern split attains comparable accuracy but uses 2.6$\times$ and 2.4$\times$ the ASN. Thus, grouping and early stopping account for much of the sampling reduction but do not reproduce the same accuracy--compute trade-off. We therefore view CG-MTE primarily as a sample-efficient adaptive inference strategy rather than a method for raising the high-budget accuracy ceiling over beam-search SC.

\begin{table}[h]
\centering
    \small
    \setlength{\tabcolsep}{3.5pt}
    \begin{tabular}{llcc}
    \toprule
    Model & Method & Acc. (\%) & ASN \\
    \midrule
    8B & SC@40 (beam) & 76.7 & 40.00 \\
     & Adaptive beam SC (rank) & 75.2 & 4.68 \\
     & Adaptive beam SC (pattern) & 75.8 & 12.72 \\
     & \textbf{CG-MTE} & 76.0 & 4.89 \\
    \midrule
    3B & SC@40 (beam) & 62.0 & 40.00 \\
     & Adaptive beam SC (rank) & 61.1 & 4.94 \\
     & Adaptive beam SC (pattern) & 62.6 & 12.28 \\
     & \textbf{CG-MTE} & 62.1 & 5.09 \\
    \bottomrule
\end{tabular}
\caption{Comparison with adaptive beam-search self-consistency baselines on PGPS9K. ASN denotes the average sample number. All adaptive methods use the same maximum budget and unique-mode stopping rule. }
\vspace{-4mm}
\label{tab:adaptive-beam}
\end{table}

\section{Conclusion}
We investigate why test-time scaling is less effective for PGP-solving and identify two key obstacles: limited reasoning diversity and perception-induced symbolic errors. 
To address these challenges, we propose Multi-Trace Synthesis, which expands symbolic programs into heterogeneous reasoning traces, and Perception-Augmented training, which grounds symbolic deduction in structured semantic clauses parsed from diagrams. 
Experiments on PGPS9K, Geometry3K, and GeoQA show that our method consistently improves geometry reasoning across model scales and achieves strong performance.
Finally, we introduce CG-MTE, a self-adaptive inference strategy that preserves most gains of high-budget self-consistency while reducing sampling cost by up to 8$\times$.

\section*{Limitations}
Despite its promising performance, our framework faces three key limitations that motivate future research. 
First, the current method depends on structured annotations. The program-seeded MTS implementation requires reliable formal solution programs, while PA training requires semantic clauses during training. This limits direct applicability to datasets without such annotations. Extending the framework to alternative reasoning seeds or less structured intermediate representations would require task-specific conversion and verification mechanisms.
Second, although we evaluate on three plane geometry benchmarks, namely PGPS9K, Geometry3K, and GeoQA, our study remains confined to 2D plane geometry. Whether the observed scaling laws and diversity-driven improvements generalize to 3D geometry, physics, or other multimodal reasoning tasks remains unverified. 
Finally, execution-based verification ensures that \CoT-\pal scripts execute successfully and return the expected answer, but it cannot verify the semantic faithfulness of all natural-language rationales. The rationales in \CoT-augmented traces may therefore occasionally be misaligned with their formal reasoning steps.

\section*{Ethics Statement}
This work uses publicly accessible geometry benchmarks, including PGPS9K, Geometry3K, and GeoQA, and does not involve private user data or sensitive personal information. Any release of derived MTS data and related artifacts complies with the licenses and redistribution terms of the source benchmarks.
The human evaluation was conducted by geometry-trained undergraduate members of the research group. Participation in this internal annotation effort was voluntary, and no separate task-specific compensation was provided. Further details on the annotation and adjudication protocol are provided in Appendix~\ref{apd:trace_quality}.
Our method is intended for research and educational use. It is not designed for safety-critical or high-stakes settings, and its outputs should not be treated as authoritative without expert verification.
More broadly, we hope this work contributes to more interpretable and verifiable AI systems. At the same time, stronger reasoning models may also produce more convincing but still incorrect outputs, underscoring the need for verification and human oversight.

\section*{Acknowledgments}
We thank all anonymous reviewers for their valuable comments. This work was supported by National Natural Science Foundation of China under No. 92370119, 62376113, 62436009, 62276258, and Jiangsu Science and Technology Programme BK20251812, and Open Research Fund of the State Key Laboratory of Multimodal Artificial Intelligence Systems.
This work was also supported by the Top Talent Reward Project under No. RDF-TP-0019.

\bibliography{custom}

\appendix

\section{Method Details}
\label{sec:method_details}

\subsection{Additional Details of the Geometric Formal Language}
\label{sec:formal_details}

We provide additional details of the domain-specific language used in Section~\ref{sec:formal_language}.

\noindent\textbf{Operators.}
The operator set $\mathcal{O}$ consists of 34 distinct geometric theorems and axioms, addressing fundamental operations involving triangles, quadrilaterals, polygons, and circles (e.g., \texttt{Gougu}).

\noindent\textbf{Operands.}
The operands $\mathsf{args}_t$ are classified into three specific categories:
(1) \textbf{Problem Variables} (\texttt{N}): Known measurements extracted from the textual problem $Q$ or semantic clauses $S$ during preprocessing and stored via an auxiliary mapping.
(2) \textbf{Process Variables} (\texttt{V}): Intermediate geometric quantities (e.g., the length of an auxiliary line) computed during the deduction process.
(3) \textbf{Constants} (\texttt{C}): Common numerical constants (e.g., $\pi$ and $180^\circ$) required for calculations.

\subsection{Multi-Trace Synthesis Details}
\label{apd:mts_details}

\subsubsection{MTS Framework Details}
\label{apd:MTS_Framework}

\begin{figure}[h]
    \centering
    \includegraphics[width=\linewidth]{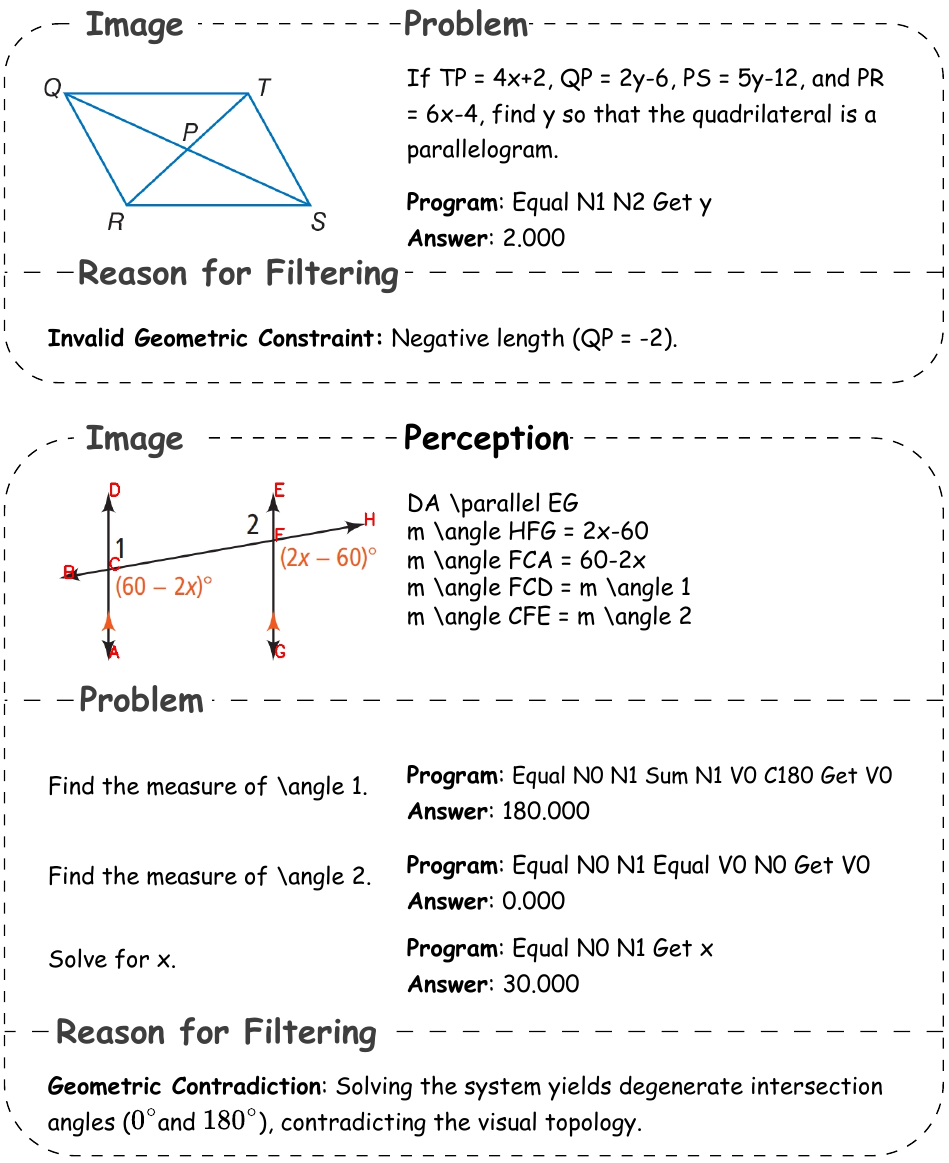}
    \caption{Examples of filtered training samples during the program-to-\pal conversion. These cases were excluded due to inherent geometric contradictions in their problem design, such as evaluating to negative segment lengths or degenerate intersection angles.}
    \label{fig:filtered_cases}
\end{figure}

To implement our MTS, we first attempt to convert the 8,021 training solution programs into executable \pal scripts. Four instances fail \pal verification because the translated solutions violate geometric constraints, such as non-negative segment-length or non-degenerate topology requirements, as illustrated in Figure~\ref{fig:filtered_cases}. These four instances are excluded only from the \pal and \CoT-\pal branches, while their original Program instances and \CoT-Program traces are retained.

This yields 8,017 verified \pal scripts and 8,017 \CoT-\pal traces, while \CoT-Program traces are constructed for all 8,021 training instances. Together, these three synthesized formats form a 24K-trace mixture named \textsc{PGPS9K-MTS}.
For \CoT-\pal generation, 7,831 traces are obtained in a single pass, 152 require a second pass, and 34 require a third pass.
We combine \textsc{PGPS9K-MTS} with the original \dataset to obtain the final training set, PGPS9K-All.

We apply the same MTS framework to Geometry3K and GeoQA, resulting in \textsc{Geometry3K-MTS} and \textsc{GeoQA-MTS}, respectively.
Table~\ref{tab:data_statistics} summarizes the statistics of the three geometry benchmarks and the resulting MTS training corpora. 

\begin{table}[h]
\centering
\small
\setlength{\tabcolsep}{6pt}
\resizebox{0.95\linewidth}{!}{
\begin{tabular}{lcccc}
\toprule
\textbf{Dataset} & \textbf{Train} & \textbf{Test} & \textbf{\# MTS} & \textbf{\# MTS-All} \\
\midrule
PGPS9K & 8,021  & 1,000 & 24.1K & 32.1K \\
Geometry3K & 8,432  & 589 & 25.3K & 33.7K \\
GeoQA & 3,485  & 754 & 10.5K & 13.9K \\
\bottomrule
\end{tabular}
}
\caption{
Statistics of geometry datasets and the resulting MTS training corpora.
``\# MTS'' denotes the number of synthesized multi-trace instances, while ``\# MTS-All'' includes both original training and synthesized traces.
}
\label{tab:data_statistics}
\end{table}

\subsubsection{Prompt Templates for Rationale Augmentation}
\label{apd:prompt_for_reasoning}

As discussed in \S\ref{sec:cot}, we utilize MLLM-based rewriting to enrich executable scripts with natural-language rationales. This section provides the exhaustive prompt templates used to generate our diverse reasoning traces.

Table~\ref{tab:prompt_cotpal} presents the prompt for \CoT-\pal transformation.
Table~\ref{tab:prompt_cotpal_fix} details the prompt for repairing \CoT-\pal, which implements a self-correction loop. When a synthesized script fails execution, the MLLM receives the traceback error as feedback to iteratively refine the code and rationales. 
Table~\ref{tab:prompt_cotprogram} presents the prompt for rewriting the program, designed to align structured geometric explanations with symbolic operator sequences. These instructions emphasize theorem-grounded justifications while maintaining the strict execution order of the original solution programs.

\section{Additional Implementation Details}
\label{apd:training_and_impl}

\subsection{Training Details}
\label{apd:implementation_details}

We implement our fine-tuning pipeline using the \texttt{SFTTrainer} from Hugging Face TRL with DeepSpeed ZeRO-2 on 8 NVIDIA A100 (80GB) GPUs.
We fine-tune three backbones: Qwen2-VL-2B-Instruct, Qwen2.5-VL-3B-Instruct, and Qwen3-VL-8B-Instruct.
We use model-specific learning rates: $2 \times 10^{-5}$ for Qwen2-VL-2B-Instruct and Qwen2.5-VL-3B-Instruct, and $5 \times 10^{-6}$ for Qwen3-VL-8B-Instruct.
All runs use 10 training epochs, a cosine learning-rate schedule with 10\% warmup, and a maximum sequence length of 1,024 tokens.
Unless otherwise stated, we use greedy decoding (\texttt{do\_sample=False} and \texttt{num\_beams=1}) for the main results reported in Table~\ref{tab:main_results}.
For test-time scaling, we evaluate deterministic beam search (\texttt{do\_sample=False}) and nucleus sampling (\texttt{do\_sample=True}) with temperature $T=0.9$ and top-$p=0.9$.

\section{Additional Experimental Analyses}
\label{apd:additional_experiments}

\subsection{Additional Results on InternVL3.5 Backbones}
\label{apd:internvl_results}

To further evaluate the robustness of our framework across different multimodal backbones, we additionally conduct experiments on InternVL3.5-2B and InternVL3.5-8B using the same training and evaluation protocols as the main experiments.

Table~\ref{tab:internvl_main} shows that PA training on PGPS9K-All consistently improves performance over direct symbolic-program prediction on both InternVL backbones, achieving gains of $+8.1$ and $+8.3$ points for the 2B and 8B models, respectively. 
Moreover, removing either PA training or PGPS9K-MTS leads to clear performance degradation, demonstrating that both components contribute consistently across model sizes. The results demonstrate that our framework consistently improves performance across different multimodal backbones.

\begin{table}[h]
\centering
\small
\resizebox{\linewidth}{!}{
\begin{tabular}{lccc}
\toprule
Setting & Training Data & 2B & 8B \\
\midrule
Baseline: Direct Program Prediction & PGPS9K & 36.7 & 40.2 \\
Ours: PA Training on PGPS9K-All & PGPS9K-All & \textbf{44.8} \improve{8.1} & \textbf{48.5} \improve{8.3} \\
w/o PA Training: Direct Prediction & PGPS9K-All & 38.8 \dec{6.0} & 42.4 \dec{6.1} \\
w/o PGPS9K-MTS: PA Training & PGPS9K & 41.7 \dec{3.1} & 45.2 \dec{3.3} \\
\bottomrule
\end{tabular}
}
\caption{Experimental results on InternVL3.5 models.}
\label{tab:internvl_main}
\end{table}

Beyond greedy decoding, we further evaluate test-time scaling on InternVL3.5 models trained with PA training on PGPS9K-All. 
As shown in Table~\ref{tab:internvl_sc}, self-consistency remains consistently effective under larger inference budgets. 
Specifically, performance improves from $44.8\%$ to $49.9\%$ on InternVL3.5-2B and from $48.5\%$ to $54.9\%$ on InternVL3.5-8B under SC@40.

\begin{table}[h]
\centering
\small
\begin{tabular}{lcc}
\toprule
Inference Budget & InternVL3.5-2B & InternVL3.5-8B \\
\midrule
Top-1 & 44.8 & 48.5 \\
SC@4  & 45.2 & 49.5 \\
SC@8  & 47.5 & 51.6 \\
SC@12 & 47.5 & 52.2 \\
SC@16 & 49.0 & 53.6 \\
SC@20 & 48.5 & 53.6 \\
SC@24 & 48.9 & 53.8 \\
SC@28 & 49.5 & 54.1 \\
SC@32 & 49.8 & 54.2 \\
SC@36 & 49.8 & 54.5 \\
SC@40 & \textbf{49.9} & \textbf{54.9} \\
\bottomrule
\end{tabular}
\caption{Test-time scaling on InternVL3.5 models under different self-consistency budgets.}
\label{tab:internvl_sc}
\end{table}

\subsection{Qualitative Analysis of PA Training}
\label{apd:perception_cases}

\begin{figure}[t]
    \centering
    \includegraphics[width=\linewidth]{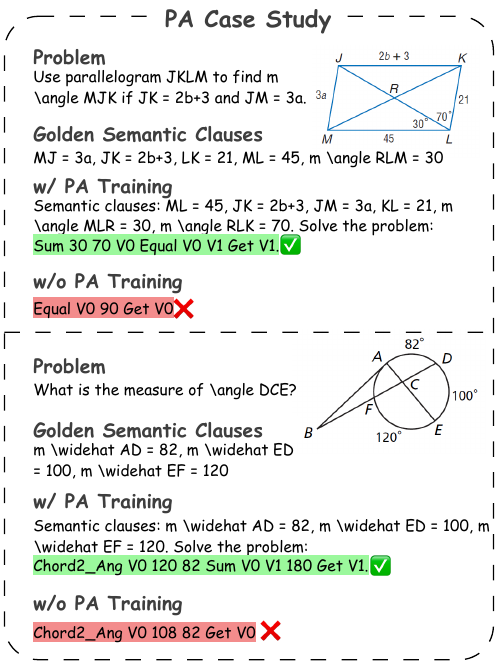}
    \caption{Qualitative comparison between direct program prediction and PA training. }
    \vspace{-4mm}
    \label{fig:perception_cases}
\end{figure}

We analyze representative cases in Figure~\ref{fig:perception_cases} to understand why PA training improves PGP-solving. 
Without PA training, the model directly maps diagram observations into symbolic operations, often producing incomplete or weakly grounded reasoning steps. 
By explicitly predicting semantic clauses before reasoning, PA training provides a more reliable visual-to-symbolic interface, reducing invalid theorem applications and hallucinated geometric steps.

\subsection{Trace Quality Assessment}
\label{apd:trace_quality}

To assess the semantic fidelity of the synthesized reasoning traces of PGPS9K, we conduct a manual evaluation on 400 randomly sampled instances from each \CoT-based format, namely \CoT-\pal and \CoT-Program. 
The evaluation was carried out by four high-performing undergraduate annotators with strong academic records and solid training in Euclidean geometry and mathematical problem solving. 
Before the formal annotation, all annotators were given a unified annotation guideline together with several calibration examples to ensure a consistent understanding of the evaluation criteria.

For each sampled trace, the annotators examine whether: 
(1) the natural-language rationale correctly reflects the underlying geometric theorem or algebraic operation; 
(2) the reasoning flow is logically consistent from premise to conclusion.
A trace is judged correct only if both criteria are satisfied. 
During annotation, the evaluators were instructed to focus on the faithfulness of the reasoning trace to the underlying symbolic process, rather than surface-level fluency alone.
Each sampled trace was independently assessed by two annotators.
When the two initial labels disagreed, a third annotator adjudicated the case by comparing the rationale with the underlying symbolic program or \pal script.

\begin{figure}[!htbp]
    \centering
    \includegraphics[width=\linewidth]{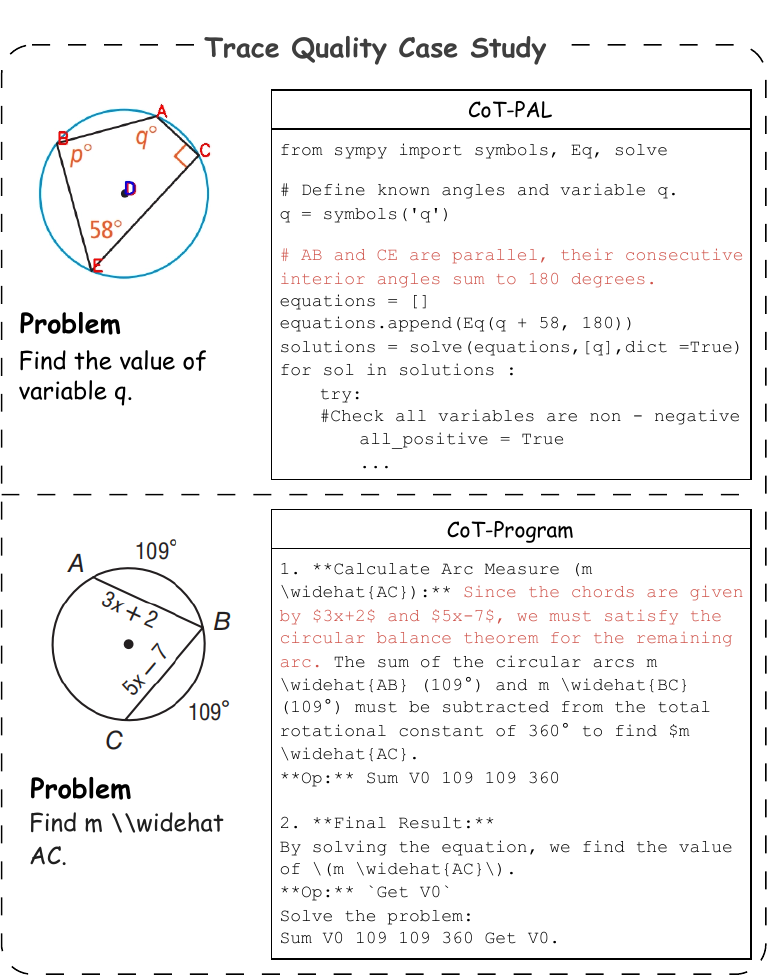}
    \caption{Representative erroneous examples from \CoT-\pal and \CoT-Program in the manual trace quality assessment. Red text highlights the logically flawed or hallucinated reasoning steps generated by the models.}
    \vspace{-4mm}
    \label{fig:trace_quality}
\end{figure}

Representative errors from both \CoT-\pal and \CoT-Program are shown in Figure~\ref{fig:trace_quality}. 
Specifically, in the \CoT-\pal example, the model hallucinates a geometric rationale that is entirely disconnected from the actual visual topology. While it correctly formulates the algebraic equation (\texttt{Eq(q + 58, 180)}), it fabricates an unverified assumption that chords AB and CE are parallel to justify the 180-degree summation via consecutive interior angles. 
In the \CoT-Program example, the generated trace fabricates a non-existent ``circular balance theorem'' and illogically attempts to justify a basic subtraction of arc measures ($360^\circ - 109^\circ - 109^\circ$) by referencing irrelevant algebraic expressions for chord lengths ($3x+2$ and $5x-7$).

\subsection{Training Efficiency of \CoT-\pal}
\label{apd:training_efficiency}
\begin{figure}[!htbp]
    \centering
    \includegraphics[width=1\linewidth]{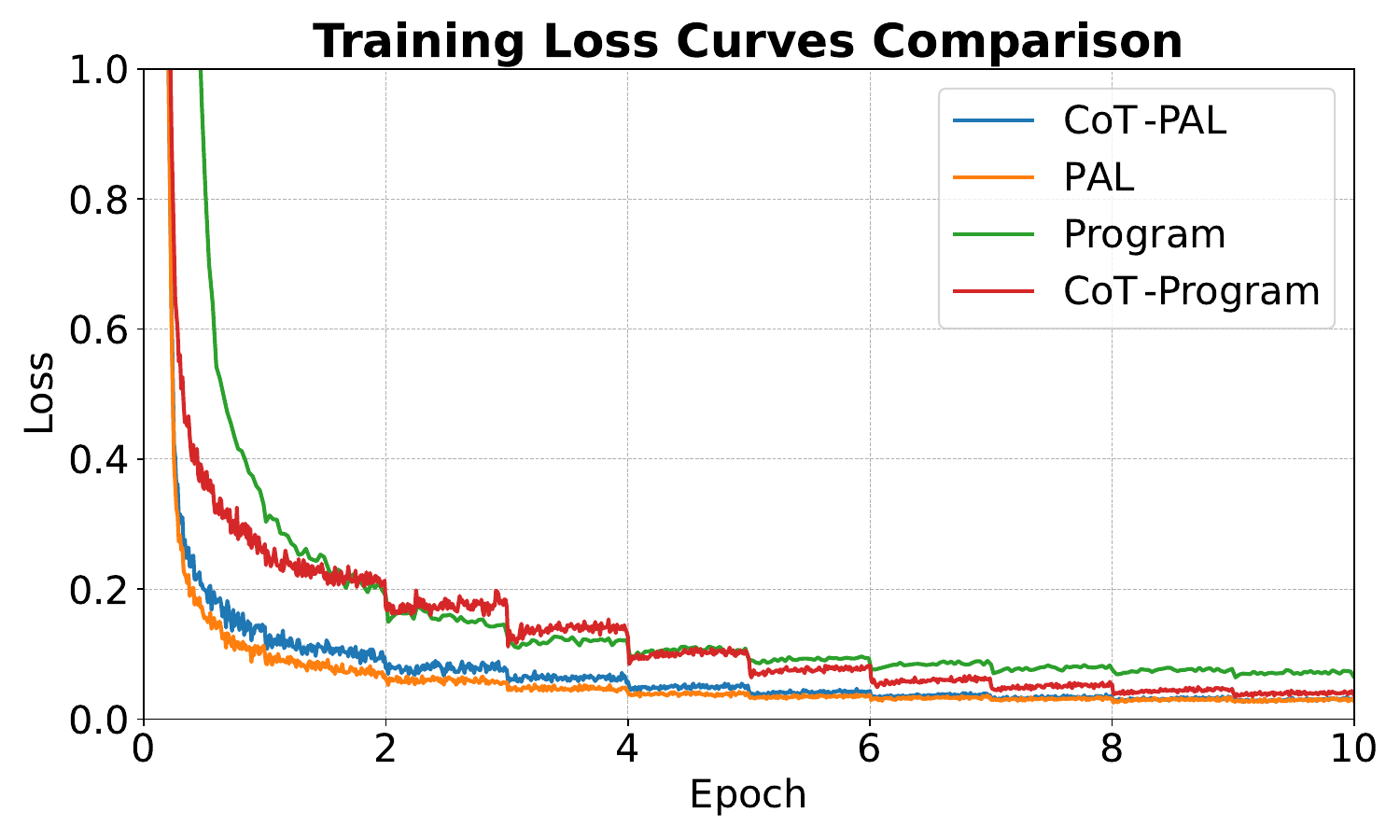} 
    \caption{Training loss trajectories on PGPS9K for single-trace models using the Qwen3-VL-8B backbone.}
    \label{fig:loss_curve}
\end{figure}

The training dynamics illustrated in Figure~\ref{fig:loss_curve} reveal that the Qwen3-VL-8B model trained on Python-based traces of PGPS9K (\pal and \CoT-\pal) exhibits accelerated convergence and achieves lower training loss than their program-based counterparts. 
One possible explanation is that Python-based reasoning trajectories are closer to the code-generation tasks seen during pre-training, making them easier for the model to learn. In contrast, the solution program contains domain-specific operators whose semantics must be learned largely from task-specific data. Consequently, \CoT-\pal provides a more learnable training target and leads to more efficient optimization during fine-tuning.

\section{Additional Analysis of Test-Time Scaling}
\label{apd:test_time_inference}

\subsection{Trace-Specific Inference Instructions}
\label{apd:mte_prompts}

For Multi-Trace Ensemble (MTE), we generate diverse reasoning trajectories by applying different inference instructions to the same fine-tuned model. 
The underlying model parameters are shared across all trace types; only the instruction suffix appended to the question is changed. Unless otherwise specified, all decoding hyperparameters remain identical across trace types.
Specifically, we use the following trace-specific inference instructions:

\begin{itemize}
    \item \textbf{Program:} 
    \texttt{\{question\} Please solve the problem using a symbolic program.}

    \item \textbf{\pal:} 
    \texttt{\{question\} Please solve the problem using Python code.}

    \item \textbf{\CoT-Program:} 
    \texttt{\{question\} Please reason step by step, then solve the problem using a symbolic program.}

    \item \textbf{\CoT-\pal:} 
    \texttt{\{question\} Please reason step by step, then solve the problem using Python code.}
\end{itemize}

\subsection{Test-Time Scaling on Three Geometry Benchmarks}
\label{apd:TTS_on_3_benchmarks}

We further evaluate test-time scaling behavior on all three geometry benchmarks using Qwen3-VL-8B trained with the MTS framework. Following prior work on self-consistency, we increase the inference budget and aggregate multiple candidate solutions during decoding. 

As shown in Table~\ref{tab:tts_three_benchmarks}, self-consistency consistently improves performance across all benchmarks. Specifically, answer accuracy improves from $71.2\%$ to $77.2\%$ on PGPS9K, from $74.5\%$ to $80.7\%$ on Geometry3K, and from $67.2\%$ to $76.1\%$ on GeoQA. These results demonstrate that models trained on diverse multi-trace reasoning data constructed by the MTS framework can effectively benefit from test-time scaling beyond PGPS9K alone. 

Notably, the improvements are particularly significant on GeoQA, where self-consistency yields an absolute gain of nearly $9\%$. We conjecture that the increased reasoning diversity introduced by MTS enables the model to generate multiple complementary solution paths, thereby improving the robustness of majority-vote decoding under larger inference budgets.

\begin{table}[t]
\centering
\small
\begin{tabular}{lccc}
\toprule
Inference Budget & PGPS9K & Geometry3K & GeoQA \\
\midrule
Top-1  & 71.2 & 74.5 & 67.2 \\
SC@4   & 72.7 & 75.4 & 73.2 \\
SC@8   & 74.9 & 77.4 & 75.3 \\
SC@12  & 76.0 & 79.6 & 74.7 \\
SC@16  & 77.1 & 79.8 & 75.2 \\
SC@20  & 76.9 & 80.3 & 75.4 \\
SC@24  & 76.9 & 80.1 & 75.6 \\
SC@28  & \textbf{77.2} & 80.5 & 75.9 \\
SC@32  & 77.0 & 80.1 & 75.6 \\
SC@36  & 76.9 & \textbf{80.7} & 76.0 \\
SC@40  & 76.7 & 80.1 & \textbf{76.1} \\
\bottomrule
\end{tabular}
\caption{Test-time scaling performance of Qwen3-VL-8B trained on MTS-All datasets under different self-consistency inference budgets.}
\label{tab:tts_three_benchmarks}

\end{table}

\subsection{Comparative Analysis of Decoding Strategies}
\label{apd:analysis_of_self_consistency}

To investigate the performance gap between decoding strategies on PGPS9K, we analyze Qwen3-VL-8B by tracking (i) top-$k$ accuracy and the distribution of the four reasoning trace types within the top-40 predictions, and (ii) the frequency of unique numerical answers. 

\begin{figure}[h]
    \centering
    \includegraphics[width=\linewidth]{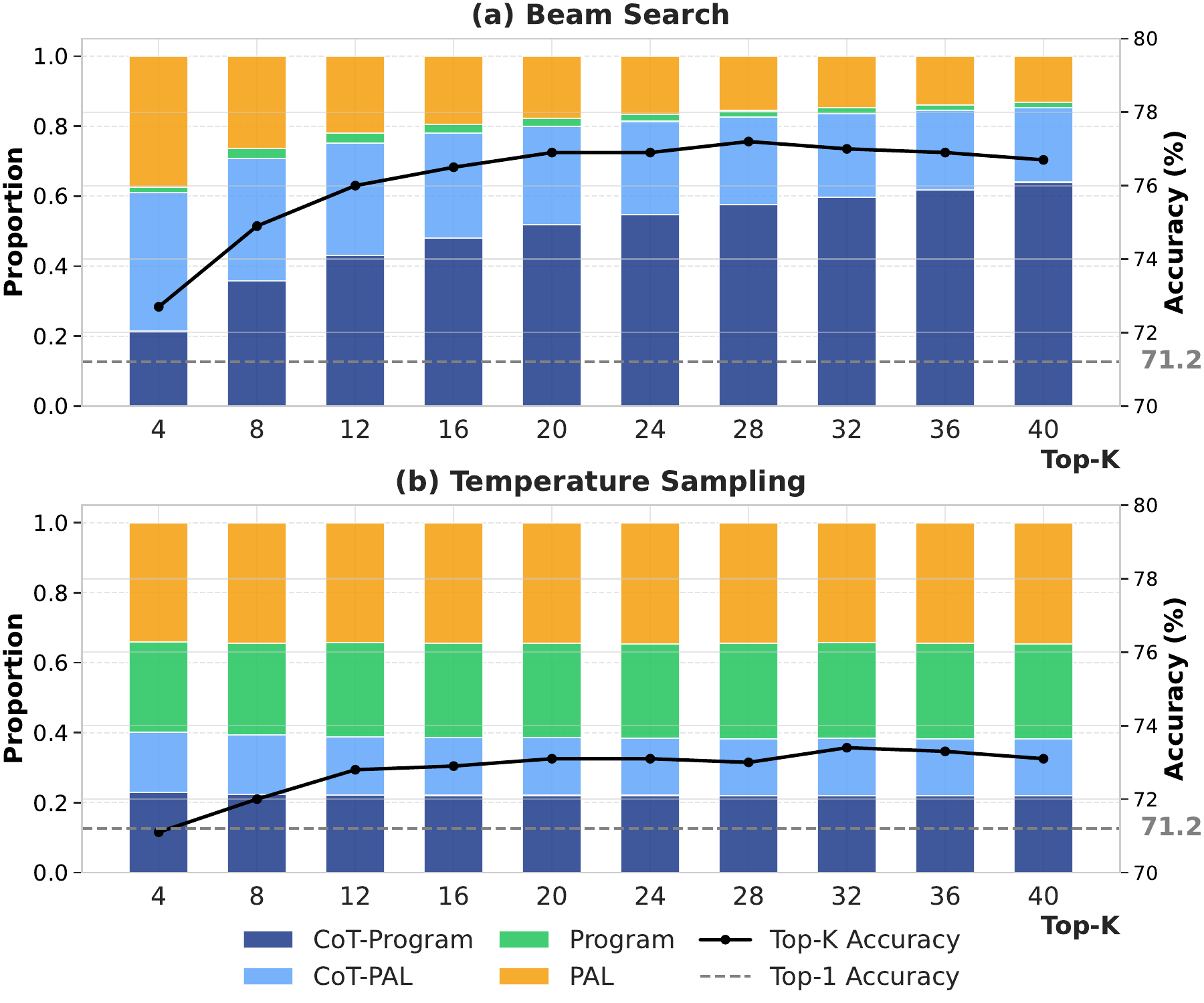}
    \caption{Evolution of reasoning trace distributions and accuracy across top-$k$ predictions for Qwen3-VL-8B on PGPS9K under (a) Beam Search and (b) Temperature Sampling.}
    \label{fig:topk_reasoning_distribution_two_sampling}
\end{figure}

\paragraph{Dynamic Reasoning Distribution Shift.}
To understand the drivers behind this scaling success on PGPS9K, we analyze the evolution of reasoning traces for Qwen3-VL-8B in Figure~\ref{fig:topk_reasoning_distribution_two_sampling} (a).
We observe a distinct hierarchical preference.
At low $k$ ($k=4$), executable traces (\pal and \CoT-\pal) dominate, accounting for $77\%$, reflecting a preference for rigorous code in high-confidence predictions.
However, as the search budget expands to $k=40$, the distribution shifts significantly: \CoT-augmented traces (\CoT-Program and \CoT-\pal) jointly constitute the vast majority ($>85\%$) of the candidates.
In contrast, Figure~\ref{fig:topk_reasoning_distribution_two_sampling} (b) shows that temperature sampling maintains a nearly static reasoning distribution across all $k$ values. The reasoning process remains concentrated on program and \pal formats, lacking explicit \CoT thought processes.

\begin{figure}[h]
    \centering
    \includegraphics[width=\linewidth]{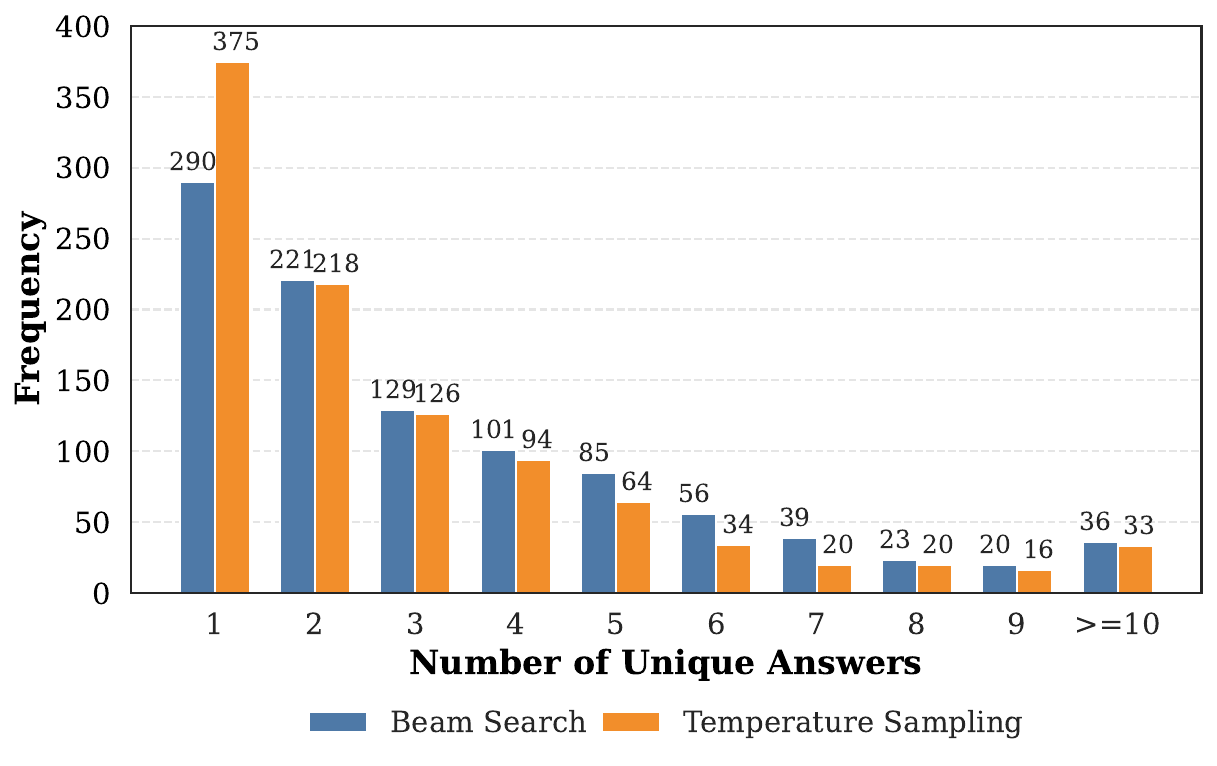}
    \caption{Distribution of unique answer counts in top-40 predictions for Qwen3-VL-8B on PGPS9K.}
    \label{fig:answer_diversity_distribution_comparison}
\end{figure}

\paragraph{Exploration Efficiency.}
We further investigate the distribution of unique numerical answers within the top-40 predictions on PGPS9K to assess the exploration efficiency. 
As illustrated in Figure~\ref{fig:answer_diversity_distribution_comparison}, temperature sampling tends to concentrate on one answer. This results in less diversity and suggests that temperature sampling may be more prone to generating repetitive or similar answers. 
In contrast, beam search generates a broader range of answers, with a higher frequency of cases involving two or more distinct answers. 
This distributional shift indicates that beam search is more effective at exploring the multiple reasoning modes.

In summary, the superiority of beam search may be partly explained by its progressive activation of \CoT-augmented reasoning as the search depth increases. This behavior fosters high-quality solution diversity, enabling majority voting to more effectively identify the correct solution from a heterogeneous pool of reasoning trajectories.

\subsection{Mechanism of Efficient Scaling}
\label{apd:consensus_guided_mechanism}

\begin{figure}[!htbp] 
    \centering
    \includegraphics[width=\linewidth]{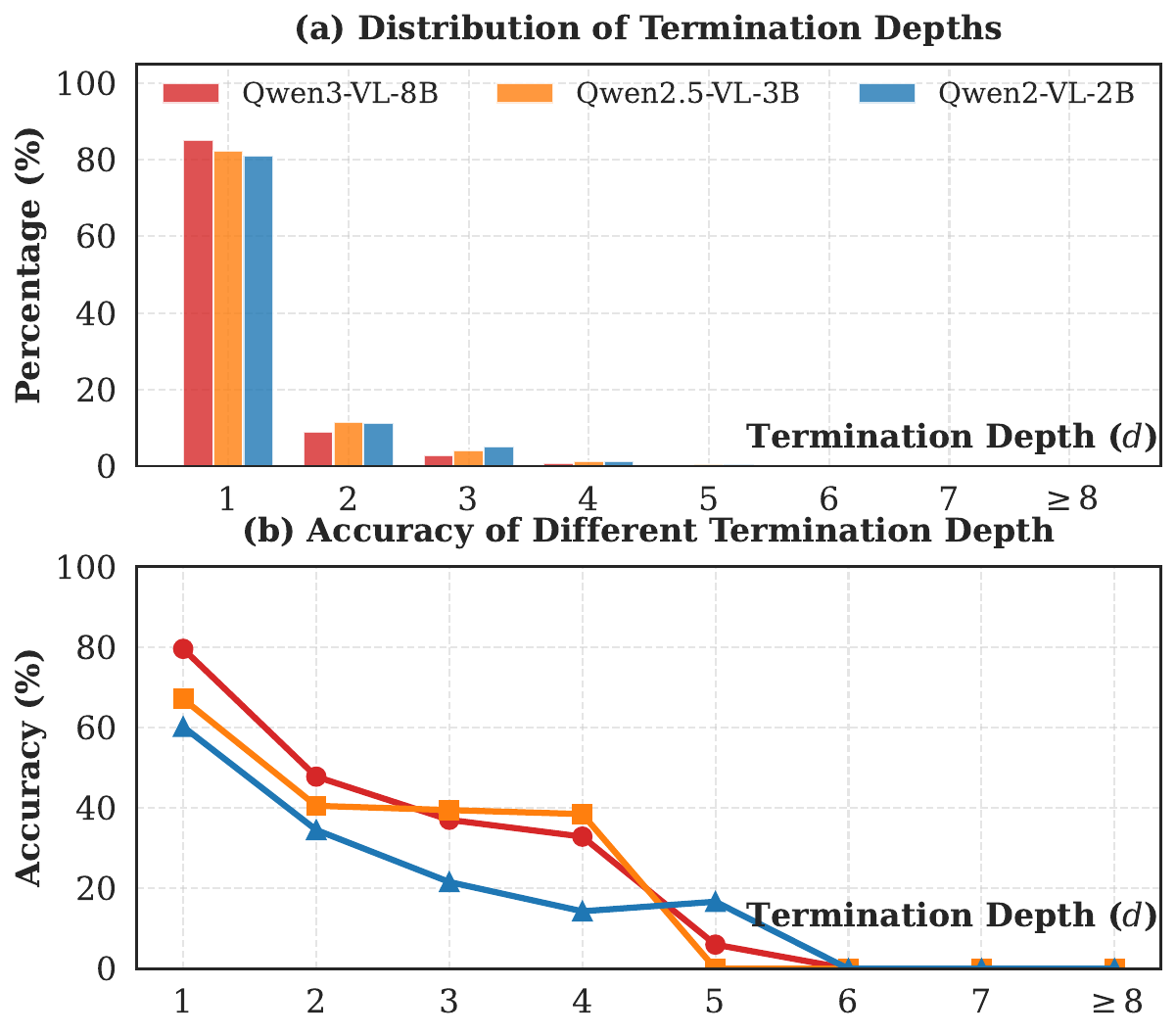}
    \caption{Mechanism of CG-MTE on PGPS9K. (\textbf{a}) Distribution of termination depths across model scales. (\textbf{b}) Model accuracy as a function of termination depth.}
    \label{fig:consensus_guided_dynamics}
\end{figure}

To explain the Pareto efficiency gains on PGPS9K, we analyze the termination behavior of the consensus-guided ensemble in Figure~\ref{fig:consensus_guided_dynamics}.
Figure~\ref{fig:consensus_guided_dynamics}(a) shows a highly skewed distribution of termination depths. 
Across all model scales, more than 80\% of problems achieve consensus immediately at $d=1$. 
This fast-track mechanism for strong inter-trace agreement problems accounts for the observed 8$\times$ reduction in sampling overhead.
Crucially, Figure~\ref{fig:consensus_guided_dynamics}(b) suggests that early consensus is associated with higher accuracy. 
Problems terminating early at $d=1$ yield the highest accuracy ($\sim$60--80\%), while deeper termination depths correspond to increased disagreement and substantially lower accuracy. 
This pattern indicates that the self-adaptive strategy effectively allocates additional computations for uncertain complex geometry problems.
Consistent with this behavior, shallow cross-trace agreement is associated with higher accuracy, suggesting that early consensus tends to occur on easier cases with more reliable predictions.

\begin{figure}[!htbp]
    \centering
    \includegraphics[width=\linewidth]{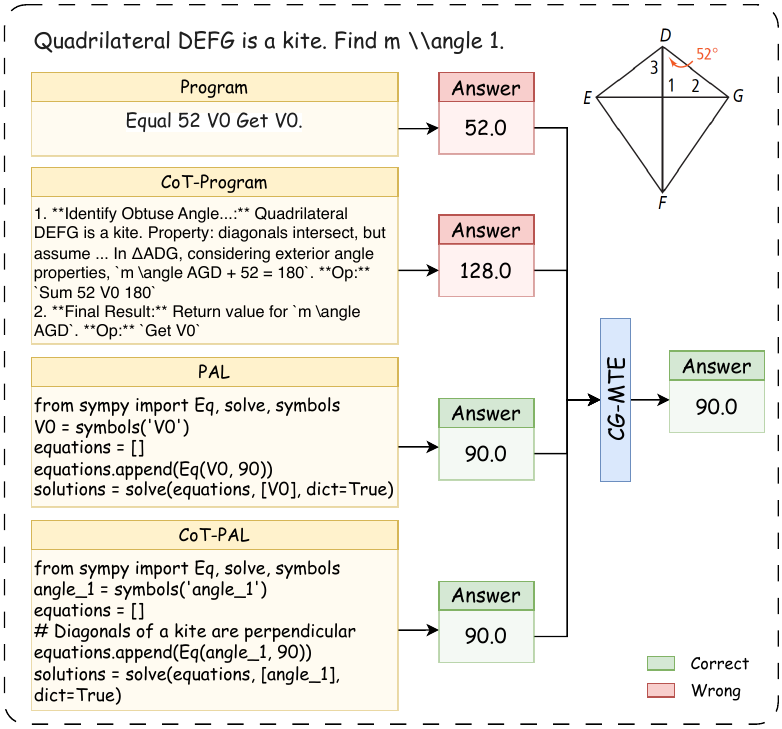}
    \caption{Representative case study of CG-MTE. A unique cross-trace mode (90.0) emerges at $d=1$, enabling immediate termination despite disagreement among individual trace formats.}
    \label{fig:consensus_guided_cases}
\end{figure}

To further illustrate how CG-MTE achieves its efficiency gain on PGPS9K, we present a representative example in Figure~\ref{fig:consensus_guided_cases}. 
This example qualitatively shows that CG-MTE reduces redundant decoding on high-consensus instances by terminating immediately once cross-trace consensus is reached at shallow depth.

\subsection{Error Coupling Across Trace Formats}
\label{sec:error-coupling}

Because the four trace outputs share the same model and program-seeded training data, their errors may not be independent.
We therefore analyze the four trace-specific top-1 outputs generated by the shared model under the Program, \pal, \CoT-Program, and \CoT-\pal prompts and used by CG-MTE on PGPS9K.
\textit{Pairwise same wrong} is computed over trace pairs for which both predictions are incorrect and measures the percentage that produce the same normalized incorrect answer. 
\textit{$\geq 3$ same wrong} and \textit{all four same wrong} denote the percentages of test instances in which at least three, or all four, trace outputs agree on the same incorrect answer, respectively.

\begin{table}[h]
    \centering
    \footnotesize
    \setlength{\tabcolsep}{3.5pt}
    \begin{tabular}{lccc}
        \toprule
        Model
        & \shortstack{Pairwise\\same wrong}
        & \shortstack{$\geq 3$\\same wrong}
        & \shortstack{All four\\identical wrong} \\
        \midrule
        2B
        & 34.3 & 14.9 & 4.9 \\
        3B
        & 34.7 & 11.5 & 3.5 \\
        8B
        & 39.8 & 8.7 & 2.5 \\
        \bottomrule
    \end{tabular}
    \caption{Error-coupling analysis among the four trace formats on PGPS9K. All values are percentages.}
    \label{tab:error-coupling}
\end{table}

As shown in Table~\ref{tab:error-coupling}, pairwise agreement on the same wrong answer ranges from 34.3\% to 39.8\%, indicating non-negligible error correlation. However, higher-order incorrect agreement is substantially less frequent: at least three traces agree on the same wrong answer in 8.7--14.9\% of test instances, while all four do so in only 2.5--4.9\%. Thus, the shared origin introduces correlated errors but does not collapse the trace formats into identical failure modes. Cross-trace consensus should therefore be viewed as an empirical early-stopping signal rather than a correctness guarantee.

\begin{table}[htbp]
    \centering
    \small
    \begin{tabular}{p{\dimexpr\linewidth-2\tabcolsep\relax}}
        \toprule
        \rowcolor{NavyBlue!7}
        \textbf{Task Definition:} \\
        You are an expert in geometry. Given (1) a geometry problem and (2) its \pal (\texttt{SymPy}) Python script. Your task is to refactor the \pal to be more human-readable by adding concise reasoning analysis before each equation. \\
        \midrule
        \rowcolor{NavyBlue!7}
        \textbf{Constraints (must follow):} \\
        1.\ Do \textbf{not} modify mathematical logic, numerical constants, or Python operators. \\
        2.\ Keep the script executable. \\
        3.\ Do \textbf{not} add, remove, or reorder any equation lines or change the final solve target. \\
        \midrule
        \rowcolor{NavyBlue!7}
        \textbf{Output format:} \\
        Output \textbf{only} the rewritten SymPy code. \\
        \midrule
        \rowcolor{NavyBlue!7}
        \textbf{In-context example:} \\
        \textbf{Input :} \\
        \begin{minipage}[t]{\linewidth}\scriptsize
        \begin{lstlisting}[style=python,  aboveskip=-4pt, belowskip=0pt]
from sympy import Eq, solve, sqrt, symbols
y, z = symbols('y z')
equations = []
equations.append(Eq(sqrt(3 * sqrt(2) * z), sqrt(2)))
equations.append(Eq(z**2 + sqrt(2)**2, y**2))
solutions = solve(equations, [y,z], dict=True)
for sol in solutions:
    try:
        # Check all variables are non-negative
        all_positive = True
        for v in sol.values():
            if float(v) <= 0:
                all_positive = False
                break
        if not all_positive:
            continue
        target_expr = y
        final_value = target_expr.subs(sol) if sol else target_expr
        val = float(final_value)
        print(f'{val:.3f}')
        break
    except (TypeError, ValueError):
        pass
        \end{lstlisting}
        \end{minipage} \\
        \textbf{Output :} \\
        \begin{minipage}[t]{\linewidth}\scriptsize
        \begin{lstlisting}[style=python,  aboveskip=-4pt, belowskip=-2pt]
from sympy import Eq, solve, sqrt, symbols
y, z = symbols('y z')
equations = []
# Geometric Mean Theorem (BD^2 = AB * BC)
equations.append(Eq(sqrt(3 * sqrt(2) * z), sqrt(2)))
# Pythagorean Theorem (in right triangle CBD: CB^2 + DB^2 = DC^2)
equations.append(Eq(z**2 + sqrt(2)**2, y**2)) 
...
        \end{lstlisting}
        \end{minipage} \\
        \bottomrule
    \end{tabular}
    \caption{Prompt for \CoT-\pal rationale augmentation.}
    \label{tab:prompt_cotpal}
\end{table}

\begin{table}[!htbp]
    \centering
    \small
    \begin{tabular}{p{\dimexpr\linewidth-2\tabcolsep\relax}}
        \toprule
        \rowcolor{NavyBlue!7}
        \textbf{Task Definition:} \\
        You are an expert Python programmer and mathematician specializing in geometry. Given: (1) the original problem (\texttt{question}), (2) the expected numeric \texttt{answer}, (3) the execution failure details (\texttt{error\_detail}), (4) the original \pal script as reference for the core symbolic logic; and (5) a buggy human-readable \CoT-\pal script with detailed natural-language comments. \\
        Your task is to produce a \emph{new}, \emph{correct}, and well-explained \CoT-\pal script that runs without errors and computes the expected \texttt{answer}. \\
        \midrule
        \rowcolor{NavyBlue!7}
        \textbf{Key requirements (must follow):} \\
        1.\ \textbf{Correctness}: the script must execute successfully and yield the provided \texttt{answer}. \\
        2.\ \textbf{Clarity}: keep (and improve if needed) the theorem-grounded natural-language comments, including diagram analysis, knowns/unknowns, principles, equation justifications, and solution process. \\
        \midrule
        \textbf{Output format:} \\
        Output \textbf{only} the corrected \CoT-\pal Python code. \\
        \bottomrule
    \end{tabular}
    \caption{Full prompt template for \CoT-\pal bug fixing.}
    \label{tab:prompt_cotpal_fix}
\end{table}

\begin{table}[!htbp]
    \centering
    \small
    \begin{tabular}{p{\dimexpr\linewidth-2\tabcolsep\relax}}
        \toprule
        \rowcolor{NavyBlue!7}
        \textbf{Task Definition:} \\
        You are an expert in Euclidean geometry and symbolic reasoning. You are given a geometry problem with:
        (1) the textual problem,
        (2) parsed geometric relations from the diagram,
        and (3) a symbolic solution program.
        Your goal is to generate a structured natural-language explanation that is \textbf{strictly aligned with each program step.} \\
        
        \midrule
        \rowcolor{NavyBlue!7}
        \textbf{Generation Guidelines:} \\
        1.\ Explicitly state the geometric meaning of each variable. \\
        2.\ Explain \emph{why} each operator is applied, including the relevant geometric theorem or principle. \\
        3.\ Keep the explanation concise and focused; avoid redundant or verbose descriptions. \\
        4.\ Do \textbf{not} change the operator sequence, operands, or final target variable. \\
        5.\ Do \textbf{not} introduce operators beyond those appearing in the original program. \\
        
        \midrule
        \rowcolor{NavyBlue!7}
        \textbf{Output Format:} \\
        Output a numbered list of reasoning steps. Each step should:
        (i) contain a short descriptive title,
        (ii) provide a brief geometric explanation, and
        (iii) explicitly indicate the corresponding operator using \texttt{Op:}. \\
        
        \midrule
        \rowcolor{NavyBlue!7}
        \textbf{In-context Example:} \\
        \textbf{Input:} \\
        \begin{minipage}[t]{\linewidth}\scriptsize
        \begin{lstlisting}[style=python, aboveskip=0pt, belowskip=0pt]
Geo_Mean 3\sqrt{2} z \sqrt{2}
Gougu z \sqrt{2} y
Get y
        \end{lstlisting}
        \end{minipage} \\
        
        \textbf{Output:} \\
        \begin{minipage}[t]{\linewidth}\scriptsize
        \begin{lstlisting}[style=python, aboveskip=0pt, belowskip=0pt]
1. **Calculate CB (z):**
   In the right-angled triangle ADC, BD is the altitude to the hypotenuse AC.
   By the geometric mean theorem, the square of BD equals the product of AB and CB.
   **Op:** `Geo_Mean 3\sqrt{2} z \sqrt{2}`

2. **Calculate DC (y):**
   Triangle BDC is right-angled at B.
   By the Pythagorean theorem, $BC^2 + BD^2 = DC^2$.
   **Op:** `Gougu z \sqrt{2} y`

3. **Final Result:**
   Return the computed value of DC.
   **Op:** `Get y`
Therefore, final program is Geo_Mean 3\sqrt{2} z \sqrt{2} Gougu z \sqrt{2} y Get y
        \end{lstlisting}
        \end{minipage} \\
        
        \bottomrule
    \end{tabular}
    \caption{Prompt template for \CoT-Program rewriting.}
    \label{tab:prompt_cotprogram}
\end{table}

\end{document}